\documentclass[runningheads]{llncs}

\usepackage{eccv}

\usepackage{eccvabbrv}

\usepackage{graphicx}
\usepackage{booktabs}

\usepackage[accsupp]{axessibility} 

\usepackage{subcaption}
\usepackage{amsmath}
\usepackage{amssymb}
\usepackage{adjustbox}
\usepackage{makecell}
\usepackage{multirow}
\usepackage{bbding}
\usepackage{color, colortbl}
\usepackage{enumitem}
\usepackage{circledsteps}

\usepackage{pifont}

\usepackage{algorithm}
\usepackage{algorithmicx}

\usepackage{algpseudocode}

\definecolor{random}{rgb}{0.54,0.39,0.66}
\definecolor{P1}{rgb}{0.91,0.74,0.21}

\definecolor{P2}{rgb}{0.24,0.73,0.58}

\usepackage{hyperref}

\usepackage{orcidlink}

\begin{document}

\title{Before Thinking, Learn to Decide: \\ Proactive Routing for Efficient Visual Reasoning} 

\titlerunning{Proactive Routing for Efficient Visual Reasoning}

\makeatletter
\def\thanks#1{\protected@xdef\@thanks{\@thanks
        \protect\footnotetext[0]{#1}}}
\makeatother

\author{
  Yinan Zhou\inst{1,2,3}\orcidlink{0009-0003-2211-6688} \and
  Haokun Lin\inst{2,3,4}\orcidlink{0009-0000-1084-7115} \and
  Yichen Wu\inst{5}\orcidlink{0000-0003-2859-3285} \and \\
  Yuxin Chen\inst{2}$^{,\dag}$\orcidlink{0000-0002-7854-1072} \and
  Teng Wang\inst{2}$^{,\dag}$\orcidlink{0000-0003-2331-3619} \and
  Caifeng Shan\inst{4}\orcidlink{0000-0002-2131-1671} \and  
  Zhenan Sun\inst{6}\orcidlink{0000-0003-4029-9935} \and \\
  Chen Ma\inst{3}$^{,\ddag}$\orcidlink{0000-0001-7933-9813} \and
  Li Zhu\inst{1}$^{,\ddag}$\orcidlink{0000-0003-2136-3196} \and
  Ying Shan\inst{2}\orcidlink{0000-0001-7673-8325}
  \thanks{Work done during internship at Tencent. $^\ddagger$Corresponding authors.
  $^\dagger$Project Lead.}
}

\authorrunning{Y. Zhou et al.}

\institute{
  Xi'an Jiaotong University \and
  ARC Lab, Tencent IEG \and
  City University of Hong Kong \and
  Institute of Automation, CAS \and
  Harvard University \and
  Nanjing University 
}

\maketitle

\begin{abstract}

Large multimodal models have achieved strong reasoning on complex visual tasks, but their inference efficiency is often restricted by long chains of thought.
A promising solution is to pair a small draft model with a large target model, enabling cooperative inference employing a routing signal that adaptively routes queries to either the draft or target model based on their difficulties for optimal efficiency and accuracy.
Yet, the remaining bottleneck is to establish a reliable query difficulty signal under multimodal settings.
Existing approaches designed for language models either rely on post-hoc token probabilities, which fall short in multimodal scenarios, 
or depend on supervised fine-tuning, which is a data-sensitive strategy. Both paradigms perform routing only after a complete output, and ignore whether the target model can actually solve the routed instances. 
To address this, we propose \textbf{PRP}, a \textbf{P}roactive \textbf{R}outing \textbf{P}aradigm that enables early decision-making by jointly evaluating the competence of both the draft and target models. 
Our \textbf{D}raft \textbf{R}ating \textbf{L}earning \textbf{(DRL)} equips the draft model with an internal confidence estimator, while \textbf{J}oint \textbf{R}ating \textbf{L}earning \textbf{(JRL)} predicts how well the target model can handle a given query, thereby prioritizing the allocation of samples it excels at rather than the hardest ones. 
These ratings enable fine-grained, instance-level \textbf{Proactive Routing} and substantially accelerate inference without compromising overall performance. Extensive experiments across multiple multimodal reasoning benchmarks validate our effectiveness and efficiency.
\keywords{Efficient MLLM Routing\and Confidence Learning \and GRPO}
\end{abstract}
    
\section{Introduction}
\label{sec:intro}

Multimodal large language models (MLLMs)\cite{liu2023visual, zhu2025internvl3, chen2024expanding, bai2025qwen2.5vl, comanici2025gemini, hurst2024gpt,11446010,11125949,10.1145/3581783.3611709,Yang_2026_CVPR,yang2025detailfusiondualbranchframeworkenhancement} have achieved remarkable performance on visual reasoning tasks. With the emergence of reinforcement learning–based post-training methods\cite{peng2025lmm,chen2025r1v,zhang2025r1vl}, their reasoning capabilities are further strengthened, exhibiting emergent behaviors such as ``aha moments'' and self-reflection. However, these powerful reasoning abilities often come with lengthy and repetitive chains of thought. Such long reasoning are unnecessary for simple problems and introduce substantial inference overhead.

\begin{figure}[t]
  \centering
  
  \includegraphics[width=1.0\linewidth]{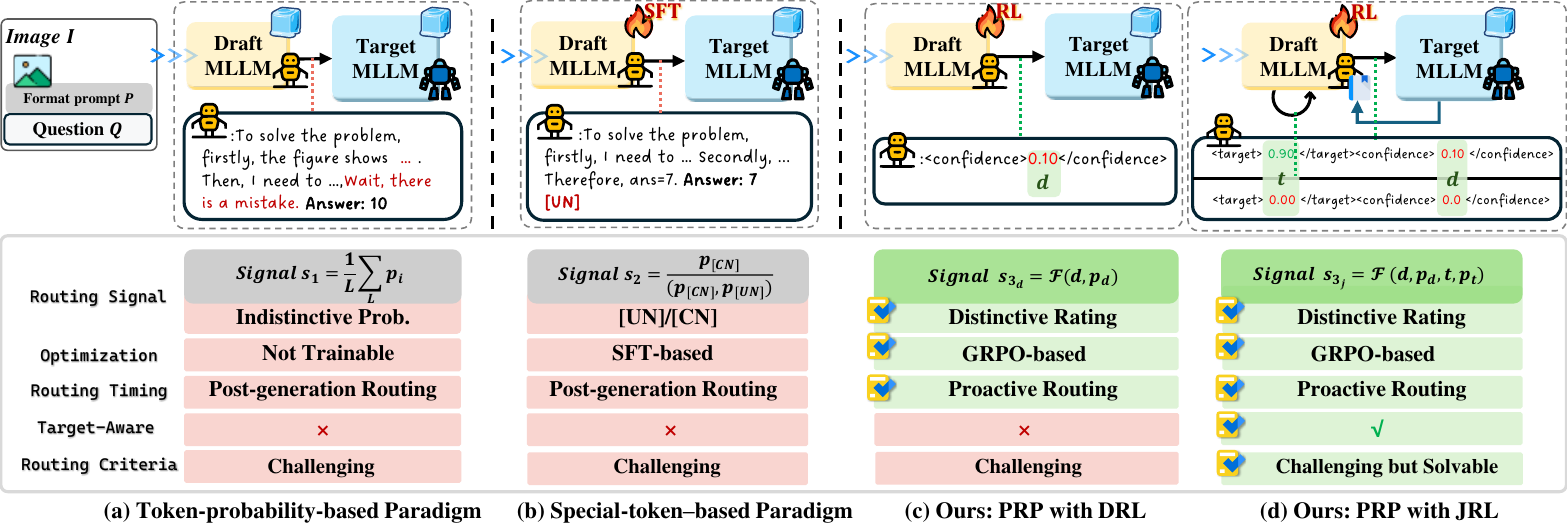}
  
   \caption{\small Overview comparing two prior paradigms with ours. We enable proactive routing at the onset of inference. Our RL-trained rater provides finer-grained and distinctive signals, and explicitly accounts for the capability of the target model.}
   \vspace{-0.2cm}
   
   \label{fig:paradigm}
\end{figure}

A natural way to reduce inference cost while preserving strong reasoning performance is to use two models of different capacities, a compact draft model that routes difficult queries to a larger target model.
Several studies explore the draft-routing paradigms for LLMs, which fall into two categories.
As illustrated in \cref{fig:paradigm}:
(i) token-probability-based routing\cite{mahaut2024factual}, which averages token probabilities as the routing signal;
(ii) special-token-based routing\cite{chuang2024self-ref} fine-tunes tokens such as $\texttt{[CN]}$ and $\texttt{[UN]}$ using correct/incorrect trajectories and routes queries based on their probabilities.

However, directly applying these routing methods to MLLMs remains under-explored. Our preliminary experiments reveal several issues.
First, token-probability-based routing provides coarse and indistinct signals, relying on an assumed positive correlation between token probability and accuracy.
Second, special-token-based routing is highly sensitive to the sampling distribution of correct and incorrect trajectories, and SFT-based methods are limited by the model’s search space.
Moreover, both paradigms defer routing decisions until the draft model finishes generation, preventing early invocation of the target model and adding latency. Importantly, they only assess difficulty from the draft model’s perspective, ignoring whether the target model is actually suitable for the instance.

To overcome these limitations, we propose a \textit{reinforcement learning-based} \textbf{P}roactive \textbf{R}outing \textbf{P}aradigm (\textbf{PRP}) for efficient visual reasoning. 
PRP initiates routing at the very beginning of generation and is aware of both the draft and target models’ capabilities.
We first introduce a fine-grained proactive rating training framework consisting of \textbf{D}raft \textbf{R}ating \textbf{L}earning (\textbf{DRL}) and \textbf{J}oint \textbf{R}ating \textbf{L}earning (\textbf{JRL}).
DRL enables the draft model to directly assess its solvability on an input before generating any reasoning steps, making proactive routing feasible.
To further evaluate the capacity of the target model, JRL additionally predicts the target model’s suitability for each instance, allowing the system to route not only difficult problems but also those that the target model handles particularly well.
We then develop a three-stage score-based \textbf{Proactive Routing} scheme for efficient MLLM inference.
The system first uses DRL or JRL trained models to produce draft and target ratings. 
These ratings are then used to rank instances using different score-based strategies. DRL model relies on fine-grained confidence distributions, while JRL model jointly evaluates the capacities of both the draft and target models to perform routing.
Our contributions are summarized as follows:

\begin{itemize}
\item[$\bullet$] We are the first to explore proactive rating for reasoning MLLMs. We propose two GRPO-based rating learning methods that enable models to assess both their own solvability and the target model’s capacity for given inputs.
\item[$\bullet$] We design several score-based inference strategies for efficient visual reasoning, jointly considering the abilities of both draft and target models.
\item[$\bullet$]
Our methods perform quick routing in the initial stage and deliver competitive reasoning performance with query-difficulty perception, while significantly reducing computation.
Notably, JRL could achieve 2.41× acceleration on MathVista while even slightly outperforming the stronger target model.
\end{itemize}

\section{Analysis of Previous Routing Paradigm}
\label{sec:preliminary}

Since previous routing methods are primarily designed for LLMs, we first conduct preliminary experiments by directly applying representative approaches to the multimodal reasoning LLMs. 
Discussed in \cref{subsec:observation}, this leads to several notable insights, further highlighting limitations in \cref{subsec:common_limitation}.

\begin{figure}[t]
  \centering

  \includegraphics[width=1.0\linewidth]{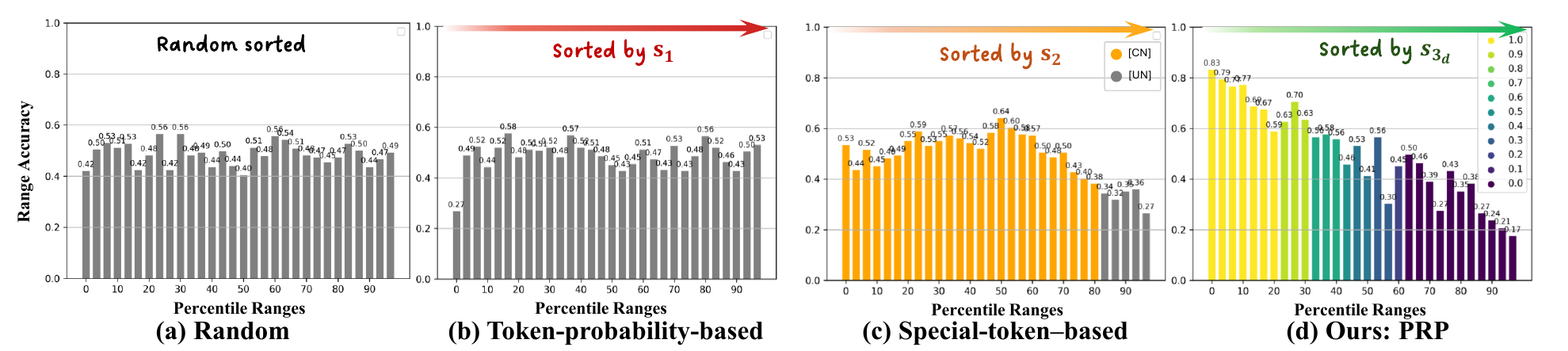}

   \caption{\small Draft model's percentile-range accuracy comparison of signal ranking across 3 paradigms versus random on MathVerse. Our PRP exhibits the performance retention and distinctive rating distribution, laying a solid foundation for fine-grained routing.}
   \label{fig:s1}
\end{figure}

\subsection{Observations}
\label{subsec:observation}

\noindent
\textbf{Token-probability-based Routing Fails in Multimodal}. 
Token-probability-based Routing Paradigm\cite{mahaut2024factual} relies on the output token-level probability distribution without training: it computes the mean probability of $s_1 = \frac{1}{L}\sum_{i=1}^{L}{p_i}$ over the generated sequence of length $L$ as a routing signal. A low $s1$ is interpreted as low confidence, triggering delegation to the target model. This strategy is coarse-grained with indistinctive signal and hinges on a presumed positive correlation between $s1$ and accuracy. However, we find that this correlation breaks down in multimodal settings with long visual reasoning: 
as shown in \cref{fig:s1}(a)(b), when sampling responses on MathVerse\cite{zhang2024mathverse} and sorting them by descending $s1$, we observe no clear monotonic relationship between $s1$ and accuracy.

\noindent
\textbf{Special-token-based Routing Paradigm Exhibits Sensitivity to SFT Data}. 
Special-token-based Routing\cite{chuang2024self-ref} starts with a pre-sampling stage before training for supervised fine-tuning data preparation. During pre-sampling, correct completions are attached by a $\texttt{[CN]}$ token and incorrect ones by a $\texttt{[UN]}$ token, enabling the representation of the correctness of its own reasoning. During training, for the correct sequence, the entire output sequence is directly fine-tuned, including the response and the $\texttt{[CN]}$ token. For the incorrect sequence, only the $\texttt{[UN]}$ token is fine-tuned to prevent performance decline. At inference time, the special token probabilities at the sequence tail provide the routing signal $s_2 = \frac{p_{[CN]}}{p_{[CN]}+p_{[UN]}}$.This approach requires supervised fine-tuning on both positive and negative trajectories with special tokens, which in turn restricts the model’s search space. Moreover, its performance is highly sensitive to the sampling distribution of positive versus negative examples in the dataset. As shown in \cref{fig:s1}(c), after trained on diverse math data MMK12\cite{meng2025mmeureka}, the draft model still struggles to reliably predict $\texttt{[CN]}$ token for the complex thinking.

\subsection{Common Limitations} 
\label{subsec:common_limitation}

Beyond the observations discussed above, these case-level routing paradigms share several common shortcomings.
First, both paradigms defer routing decisions until the draft model has generated the entire response. As a result, the target model cannot be invoked earlier, which restricts potential efficiency gains and introduces unnecessary latency.
Second, they fail to evaluate whether the selected target model is actually suitable for the given query. 
For instances that exceed the target model’s capability, routing them to the target model leads only to repetitive or low-value reasoning, offering limited performance improvement while constraining inference speed.
These limitations motivate our proactive routing paradigm for efficient visual reasoning.

\section{Training Method}
\label{sec:training_method}

\subsection{Preliminary} 
{\bf Group Relative Policy Optimization (GRPO).}
As a post-training method, GRPO improves reasoning by comparing multiple completions per sample, scoring format compliance and answer accuracy, and using the reward differences to compute advantages. In the multimodal setting, the policy model $\pi_{\theta}$ and reference model $\pi_{\theta_{old}}$ are initialized with pretrained weights. Given an input $x=(P,I,Q)$, where $P$ is a predefined formatting prompt and $(I,Q)$ is an image-question pair from dataset $\mathcal{D}$, the reference model $\pi_{\theta_{\mathrm{old}}}$ samples $N$ candidate outputs $\mathcal{O}=\{o_1,o_2,\ldots,o_N\}$. For each candidate $o_i$, we compute a format reward $R_{{fmt}_i}$ and an accuracy reward
$R_{{acc}_i}$, and define the total reward:
$ R_{{base}_i} = R_{{fmt}_i} + R_{{acc}_i}$. Within each group of $N$ candidates, we estimate the advantage of $o_i$ relative to the group average,

\begin{equation}
\hat{A}_{\text{base}_i} = \frac{R_{\text{base}_i} - \text{mean}\big(\{R_{\text{base}_j}\}_{j=1}^{N}\big)}{\text{std}\big(\{ R_{\text{base}_j} \}_{j=1}^{N}\big)},
\end{equation}

Here, the mean and standard deviation are computed over the $N$ candidates in the same group. 

Given these advantages, we apply a single-update loss similar to standard GRPO without KL penalty or clipping:

\begin{equation}
\mathcal{L}_{\text{base}_i}(\theta)=-\mathbb{E}_m\left[\frac{\pi_\theta(o_{i,m}|x,o_{i,<m})}{\pi_{\theta_{\text{old}}}(o_{i,m}|x,o_{i,<m})} \right]  \hat{A}_{\text{base}_i},
\end{equation}
where $o_{i,<m}$ is the prefix up to token $m$. Building on this objective, we extend GRPO to endow the model with confidence estimation for multimodal inputs before producing long responses, which supports both self-evaluation and cross-model evaluation.

\begin{figure*}[t]
  \centering

  \includegraphics[width=1.0\linewidth]{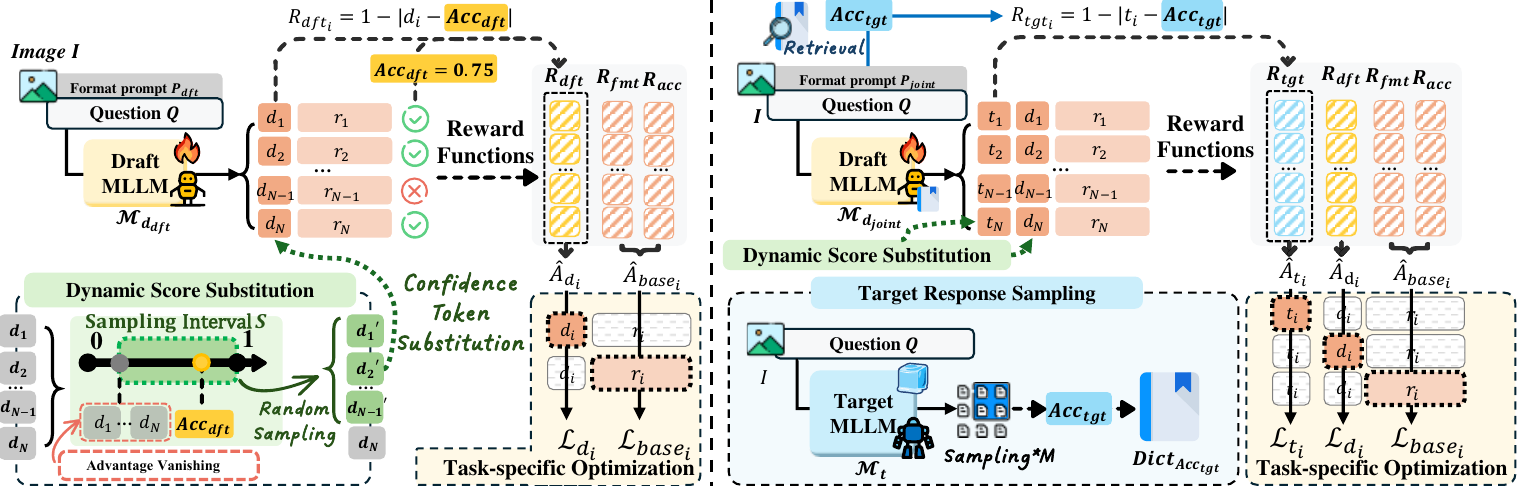}

   \caption{\small Left: Illustration of Draft Rating Learning. We design reward $R_{dft}$ and adopt Dynamic Score Substitution to avoid advantage vanishing, while Task-specific Optimization independently optimizes the $R_{dft}$-related tokens during training. Right: Illustration of Joint Rating Learning with the target model. We extend the Draft Rating Learning pipeline by introducing $R_{tgt}$, which is trained with pre-sampled ${Acc_{tgt}}$ as supervision, thereby accounting for the capability of the target model.}

   \label{fig:confidence_learning}
\end{figure*}

\subsection{Draft Rating Learning (DRL)}
\label{sec:dft_learning}
As shown in \cref{fig:confidence_learning}(left), we introduce a formatted prompt $P_{dft}$ that instructs the draft model $\mathcal{M}_{d_{dft}}$ to produce a draft score before the think phase begins. This score reflects the model's self-assessed confidence for the input $x$. The output  of $\mathcal{M}_{d_{dft}}$, denoted $o_i$, contains two parts: $d_i$, the draft score, and $r_i$, the remaining segment that includes the format markers, the reasoning, and the final answer. To train this capability effectively, we define a reward for draft rating learning, adopt a dynamic score substitution strategy to prevent advantage vanishing, and apply task-specific optimization to preserve overall performance.

\noindent{\bf Draft Rating Reward.} To train the draft model to rate both the input $x$ and its own capability, we define the Draft Rating Reward. Given $N$ sampled outputs $\mathcal{O}=\{o_i\}_{i=1}^N$, we first estimate the accuracy on this set. Specifically, we compute the mean draft accuracy as,
\begin{equation}
 \text{Acc}_{\text{dft}}=\frac{1}{N}\sum_{i=1}^{N}\mathbf{1}\{\text{correct}(o_i,\mathcal{A})\},
\label{eq:accdft}
\end{equation}
where $correct(o_i, \mathcal{A})$ denotes that the answer extracted from $o_i$ is consistent with the answer $\mathcal{A}$ of $(I, Q)$.

We then utilize $Acc_{dft}$ as the ground-truth label for the draft rating $d_i$. In this way, for inputs that are simple and within the model's competence, $d_i$ should approach 1; conversely, for inputs that are difficult or consistently produce incorrect solutions, $d_i$ should approach 0. Accordingly, we define a simple yet effective draft rating reward: 
\begin{equation}
R_{\text{dft}_i} = 
\begin{cases} 
   1 - \left| d_i - \text{Acc}_{\text{dft}} \right|, & \text{if } 0 \leqslant d_i \leqslant 1 \\
   -1, & \text{otherwise}
\end{cases}
\label{eq:rdft}
\end{equation}
To keep rewards comparable across samples and to prioritize cases with larger score discrepancies, we compute the advantage $\hat{A}_{d_i}$ by mean-centering without standard-deviation scaling,
\begin{equation}
\hat{A}_{d_i} = R_{\text{dft}_i} - \frac{1}{N} \sum_{j=1}^{N} R_{\text{dft}_j}
\label{eq:adft}
\end{equation}

\noindent{\bf Dynamic Score Substitution.} During training, we observe that the draft scores $\{d_1,\ldots,d_N\}$ produced by the model often collapse to nearly identical values. As a result, both the reward and advantage diminish, and the model stops learning a meaningful rating. We define {\textit{Advantage Vanishing}} to occur when $|mean(\{d_i\}_{i=1}^N)-Acc_{dft}|>\mu$ and $std(\{d_i\}_{i=1}^N)<\sigma$, where $\mu$ and $\sigma$ are hyperparameters. In this regime, the model tends to overuse locally high-probability tokens when predicting scores. To counteract this, we propose Dynamic Score Substitution. Specifically, we construct a sampling interval $\mathcal{S}$ as,
\begin{align*}
\mathcal{S} \in [ & \max(0,\min(d_N,\text{Acc}_{\text{dft}}) - \big|\text{Acc}_{\text{dft}} - d_N\big|), \\
& \min(\max(d_N,\text{Acc}_{\text{dft}}) + \big|\text{Acc}_{\text{dft}} - d_N\big|, 1)].
\end{align*}

\noindent Within $\mathcal{S}$, we randomly sample $\{d'_1, \ldots, d'_{N\!-\!1}\}$ to substitute $\{d_1, \ldots, d_{N\!-\!1}\}$ in $\mathcal{O}$. By construction, each ${d_{i}^{'}}$ lies closer to $Acc_{dft}$ than $d_i$, thereby yielding a higher reward $R_{dft_i}^{'}$. To amplify the advantage of the substituted scores while reducing the advantage of $d_N$, we replace $\hat{A}_{d_i}$ with $\hat{A}_{d_i}^{'}$ as

\begin{equation}
\hat{A}_{d_i}' = 
\begin{cases} 
   R_{\text{dft}_i}' - R_{\text{dft}_N}, & i \in \{1, \dots, N-1\} \\
  -\sum_{j=1}^{N-1}\hat{A}_{d_j}', & i = N
\end{cases}
\end{equation}

\noindent{\bf Task-specific Optimization.} Since $\hat{A}_{base_i}$ is to optimize format and correctness, and $\hat{A}_{d_i}$ is to accurately assess the existing model's capabilities, directly mixing $\hat{A}_{base_i}$ and $\hat{A}_{d_i}$ to optimize all output tokens $o_i$ can lead to damage to the original performance and inefficiency in rating learning. Therefore, we propose Task-specific Optimization. Specifically, we optimize the corresponding token parts, $d_i$ and $r_i$, using $\hat{A}_{d_i}$ and $\hat{A}_{base_i}$, respectively, with weight $\alpha$: 

\begin{align}
\mathcal{L}_{\text{dft}_i}(\theta) &= \mathcal{L}_{\text{base}_i}(\theta)+\alpha\mathcal{L}_{d_i}(\theta)
\notag\\=&
-\mathbb{E}_m
\frac{\pi_\theta(r_{i,m}|x,d_{i,<m},r_{i,<m})}{\pi_{\theta_{\text{old}}}(r_{i,m}|x,d_{i,<m},r_{i,<m})}  \hat{A}_{\text{base}_i}
\notag\\&-\alpha\mathbb{E}_m\frac{\pi_\theta(d_{i,m}|x,d_{i,<m},r_{i,<m})}{\pi_{\theta_{\text{old}}}(d_{i,m}|x,d_{i,<m},r_{i,<m})}  \hat{A}_{d_i}.
\end{align}

\subsection{Joint Rating Learning with Target model (JRL)} 
\label{sec:joint_learning}

As shown in  \cref{fig:confidence_learning}(right), $P_{joint}$ requires the model to output a target score $t_i$ prior to draft score $d_i$ and think phase, which denotes the target model’s assessment of the input $x$. Different from \cref{eq:accdft}, which calculates $Acc_{dft}$ during training, we perform $M$ sampling iterations on each sample in $\mathcal{D}$ using the target model before training the draft model, and record the average accuracy of $M$ outputs in $Dict_{Acc_{tgt}}$.  During training, we retrieve the corresponding $Acc_{tgt}$ as the ground-truth label of $t_i$. We apply formulas similar to \cref{eq:rdft} and \cref{eq:adft} to obtain $R_{tgt_i}$ and $\hat{A}_{t_i}$, and employ Dynamic Score Substitution to avoid advantage vanishing, along with task-specific optimization with weight $\beta$ for target rating training:

\begin{align}
\mathcal{L}_{\text{joint}_i}(\theta) &= \mathcal{L}_{\text{base}_i}(\theta)+\alpha\mathcal{L}_{d_i}(\theta)+\beta\mathcal{L}_{t_i}(\theta)
\notag\\=&
-\mathbb{E}_m
\frac{\pi_\theta(r_{i,m}|x,t_{i,<m},d_{i,<m},r_{i,<m})}{\pi_{\theta_{\text{old}}}(r_{i,m}|x,t_{i,<m},d_{i,<m},r_{i,<m})}  \hat{A}_{\text{base}_i}
\notag\\&-\alpha\mathbb{E}_m\frac{\pi_\theta(d_{i,m}|x,t_{i,<m},d_{i,<m},r_{i,<m})}{\pi_{\theta_{\text{old}}}(d_{i,m}|x,t_{i,<m},d_{i,<m},r_{i,<m})}  \hat{A}_{d_i}
\notag\\&-\beta\mathbb{E}_m\frac{\pi_\theta(t_{i,m}|x,t_{i,<m},d_{i,<m},r_{i,<m})}{\pi_{\theta_{\text{old}}}(t_{i,m}|x,t_{i,<m},d_{i,<m},r_{i,<m})}  \hat{A}_{t_i}.
\end{align}

\section{Proactive Routing Method}
\label{sec:infer_method}
\begin{figure}[t]
  \centering
  \includegraphics[width=1.0\linewidth]{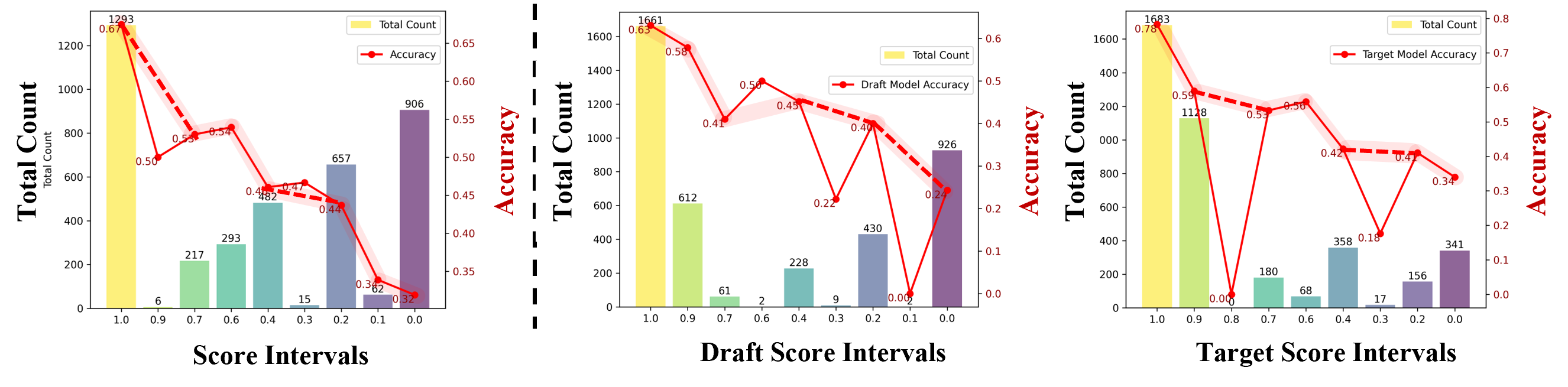}
   \caption{\small Left: The per-score instance count and accuracy of $\mathcal{M}_{d_{dft}}$ on MathVerse. Right-1: $\mathcal{M}_{d_{joint}}$’s draft score distribution on MathVerse and the draft’s corresponding accuracy. Right-2: The target score distribution and $\mathcal{M}_{t}$’s corresponding accuracy. In all plots, bars denote the number of samples in each score bin, and lines indicate the accuracy within each bin.}
   \label{fig:score_stat}
\end{figure}

\begin{figure}[t]
  \centering
  \includegraphics[width=1.0\linewidth]{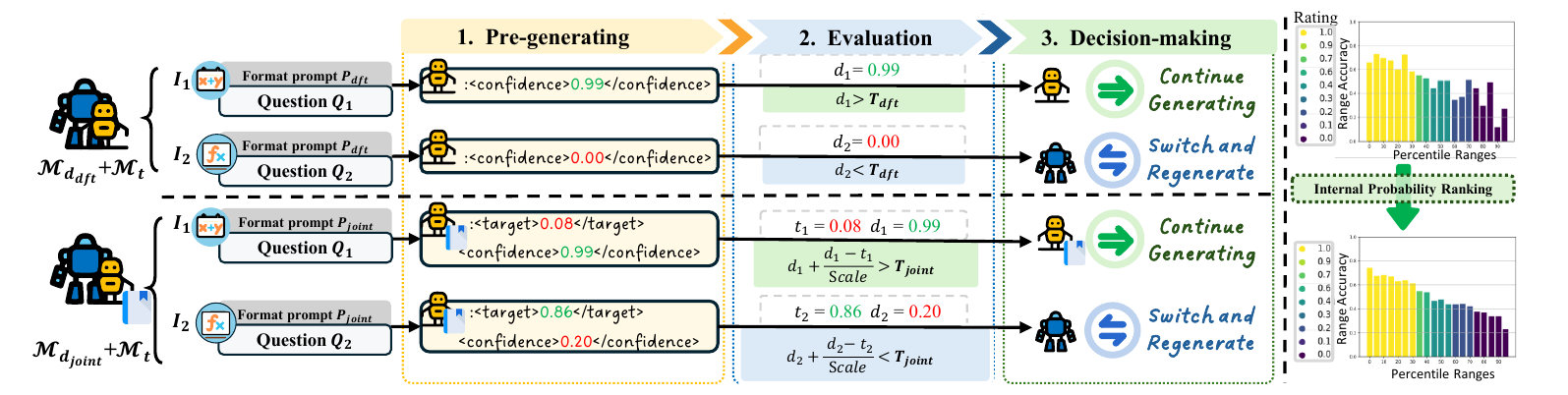}
   \caption{\small Left: Three-stage Score-based Proactive Routing. Right: Internal Probability Ranking exhibits a strong correlation with accuracy.}
   
   \label{fig:confidence_inference}
\end{figure}
In this section, we first establish the correlation between the predicted ratings and accuracy. Building on this, we then design a three-stage proactive routing scheme that delivers efficient acceleration while preserving performance. Moreover, we introduce a fine-grained partitioning method based on the token-level probabilities of predicted ratings, thereby enabling a more precise evaluation stage. We further analyze the superiority and practicality of our routing method.
\subsection{Correlation between Ratings and Accuracy}
\label{sec:correlation}
As shown in \cref{fig:score_stat}, we analyze the ratings produced by $\mathcal{M}_{d_{dft}}$ (trained in \cref{sec:dft_learning}) and $\mathcal{M}_{d_{joint}}$ (trained in \cref{sec:joint_learning}) on  MathVerse\cite{zhang2024mathverse}. The left panel reports the relationship between the ratings from $\mathcal{M}_{d_{dft}}$ during inference and accuracy. The two right panels, under $\mathcal{M}_{d_{joint}}$, respectively show (i) the draft model's score bins and its corresponding accuracies, and (ii) $\mathcal{M}_{t}$’s score bins and the target model's accuracies. Across all settings, we observe a clear trend: higher score bins correspond to higher correctness. Minor deviations in a few bins arise primarily from limited sample counts within those intervals. Overall, the yellow bins exhibit substantially higher accuracy than the purple bins, demonstrating that our approach learns meaningful, discriminative ratings that correlate strongly with accuracy.  We further demonstrate the advantages of our ratings by the correlation of interval accuracy and ratings in \cref{tab:metric}.

\subsection{Three-stage Score-based Proactive Routing}
\label{sec:routing}
Our Proactive Routing strategy involves $\mathcal{M}_{d_{dft}}$ and $\mathcal{M}_{d_{joint}}$ as routing models and follows the rating ranking in \cref{sec:correlation}. It proceeds in three stages:

\begin{itemize}
    \item {\bf Pre-generating:} Before the model starts thinking and answering, we prompt $\mathcal{M}_{d_{dft}}$ or $\mathcal{M}_{d_{joint}}$ to generate a short prefix until a complete rating snippet is obtained. From this snippet, we extract: (i) the self-assessed confidence $d$ of $\mathcal{M}_{d_{dft}}$; (ii) both the confidence $d$ of $\mathcal{M}_{d_{joint}}$ and the predicted confidence $t$ of $\mathcal{M}_t$.
    \item {\bf Evaluation:} When $\mathcal{M}_{d_{dft}}$ serves as the routing model, we directly employ $d$ as the score. When $\mathcal{M}_{d_{joint}}$ serves as the routing model, we combine $d$ and $t$ to reflect two priorities: (i) routing instances that the draft model can solve correctly to the draft model; (ii) preferring the draft model when $\mathcal{M}_{t}$ is likely to fail due to capability mismatch.
    Concretely, 
    for datasets where easy instances can be solved with high confidence, we adopt a draft-rating-first scoring strategy rule and set $s = d + \frac{d - t}{\tau}$; whereas for more challenging datasets-where the system has generally low confidence, we adopt target-rating-first scoring rule and set $s = t + \frac{t - d}{\tau}$, where $\tau$ controls the priority given to capability gap between $\mathcal{M}_{d_{joint}}$ and $\mathcal{M}_{t}$.
    \item {\bf Decision-making:} With $\mathcal{M}_{d_{dft}}$ as the router,
    if $d >T_{dft}$, we route the instance to $\mathcal{M}_{d_{dft}}$ to continue generating a full response; otherwise, we route it to $\mathcal{M}_{t}$. 
    With $\mathcal{M}_{d_{joint}}$ for routing, challenging but solvable samples have lower scores compared to overly difficult ones, which means we prioritize assigning more suitable hard samples to the target model.If {\small $s\!\!>\!\!T_{joint}$}, {\small $\mathcal{M}_{d_{joint}}$} is confident, and we route to {\small $\mathcal{M}_{d_{joint}}$} and continue generating. If {\small $s\!\!<\!\!T_{joint}$}, {\small $\mathcal{M}_{d_{joint}}$} is uncertain with lower expected accuracy, and we route to {\small $\mathcal{M}_{t}$}. 
\end{itemize}

\subsection{Internal Probability Ranking}

Relying solely on the rating score to partition samples is too coarse; many examples cluster within the same score band, leaving no principled way to prioritize them. As shown in the upper-right of \cref{fig:confidence_inference}, we first sort MathVerse samples by score and measure range accuracy across percentiles. Although the high-score (yellow) region covers a large proportion, its internal accuracy is highly heterogeneous. To resolve this, we propose Internal Probability Ranking, which refines prioritization within the same score band by using the logits of the rating tokens.

Concretely, considering a score such as ``0.98'', we examine the token position corresponding to ``9'' in the rating string and extract its logits vector $z$ over the full vocabulary.
Let $D = \{0,1,…,9\}$ denote the digit-token set, and let $p_i$ be the probability assigned to digit $i$ under the full-vocabulary softmax at that position, i.e., $p_i = \mathrm{softmax}(z)_i$ for $i\in D$. Let $p_{max}=max_{i\in D} p_i$.  We then define the within-band priority score as:

\begin{equation}
p =\\
\left\{\begin{matrix} 
   \frac{1}{|D|} p_{max}\sum_{i \in D} (p_i - \frac{1}{|D|}\sum_{j \in D} p_j)^2, \text{if } d >5 \\
  -\frac{1}{|D|} p_{max}\sum_{i \in D} (p_i - \frac{1}{|D|}\sum_{j \in D} p_j)^2,  \text{else}
\end{matrix}\right.
\end{equation}
Intuitively, when scores are high, a more concentrated digit distribution together with a larger $p_{max}$ indicates greater certainty, which yields a larger $p$ and correlates with higher accuracy; for low scores, the trend reverses.
Finally, we sort samples within each score band by $p$. As shown in the lower-right of \cref{fig:confidence_inference}, the interal accuracy then exhibits a clear stepwise pattern, which enables precise and flexible routing. We implement this by augmenting the ratings with their within-band priorities, setting ${d' = d + p_d}$ and $t' = t + p_t$. Replacing the second-stage evaluation rating $d$ and $t$ with $d'$ and $t'$ yields finer-grained assessments and more reliable routing within score ranges.

\begin{figure*}[t]
  \centering
    \vspace{-0.3cm}
  \includegraphics[width=1.0\linewidth]{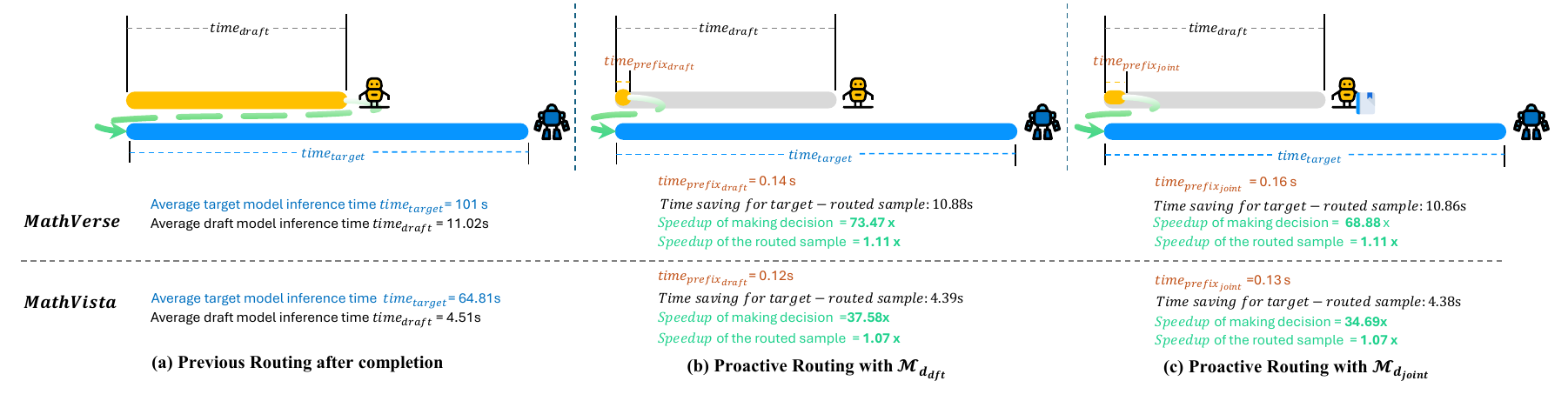}
    \vspace{-0.3cm}
   \caption{Comparison of decision latency between traditional routing-after-completion and our proactive routing schemes ($\mathcal{M}_{d_{dft}}$ and $\mathcal{M}_{d_{joint}}$).}
   \label{fig:time_analysis}
   \vspace{-0.4cm}
\end{figure*}

\subsection{Superiority and Practicality of PRP} 
\cref{fig:time_analysis} compares the end-to-end latency of routing schemes on routed-to-target cases. In the conventional ``route-after-completion'' setting, the total time is $time_{draft}+time_{target}$. With proactive routing, the draft model only generates a short prefix before making a routing decision, so the total time becomes $time_{prefix_{draft}}+time_{target}$. This saves $time_{{draft}}-time_{prefix_{draft}}$ per routed example.
On both MathVista and MathVerse, both DRL and JRL save nearly the entire inference time of the draft model, thereby advancing the routing decision to the initial stage of inference. This enables the swift redirection of challenging inputs to a more powerful target model, meaning our decision lead time is dozens of times earlier compared to post-inference routing. 
We provide two easy strategies for threshold setting as examples to illustrate the practicality of the routing method:
{\textit{(a) Task-Specific Calibration with Validation Set:}} For specific domains with a validation set, such as mathematical reasoning in MathVista, we use the validation set routing curve to find the point that achieves the best trade-off between performance and efficiency. Then, we can obtain the rating of the sample at that point. We route samples based on this rating as a threshold to achieve similar performance and speedup on test data. In \cref{fig:ablations}(f), we use 30\% of MathVista for validation and 70\% for testing. Similar trends in the curves indicate the effectiveness of this strategy.
{\textit{(b) Dynamic Pre-scoring without Validation Set:}} In the absence of a validation set, we process concurrent requests as a batch at the same time. For query batches, we perform a rapid proactive rating (taking approximately 1\% of the total draft latency shown in \cref{fig:time_analysis}). By sorting the ratings within a batch, we determine the switching threshold based on limited computational resources, allowing us to dynamically allocate these examples to maximize performance within the constraints of available resources.

\section{Experiments}
\label{sec:experiments}

\subsection{Setup}

\textbf{Evaluation Tasks.}
We evaluate the effectiveness of PRP across two representative domains. We first validate it in a simple question-answering setting, using 15K ChartQA\cite{masry2022chartqa} training subset for training and 2.5K test data for in-domain evaluation.
The second domain targets mathematical reasoning: we simply employ 15K MMK12\cite{meng2025mmeureka} for confidence learning. We adopt two widely used benchmarks that span a broad range of problem types and difficulty levels, MathVista\cite{lu2024mathvista} and MathVerse\cite{zhang2024mathverse}. We also provide the results on M3CoT\cite{m3cot} in \cref{sec:m3cot} to further illustrate our effectiveness and generalization.

\noindent
\textbf{Reasoning MLLMs.}
For QA domain, we directly train Qwen2.5-VL-3B\cite{bai2025qwen2.5vl} with DRL to obtain $M_{d_{dft}}$. We use Ovis-2.5-9B\cite{lu2025ovis2.5} as the target model.
For math domain, we first perform GRPO to empower Qwen2.5-VL-7B\cite{bai2025qwen2.5vl} with strong reasoning ability, then perform DRL and JRL on the GRPO-trained model for better efficiency, while Skywork-R1V-38B\cite{shen2025skyworkr1v3technicalreport} is adopted as $\mathcal{M}_t$.

\noindent
\textbf{Implementation Details.} $\alpha$ and $\beta$ are set to $1.5$. Rollout $N$ is set to $4$ and sample number $M$ for JRL is $16$. $M_{d_{dft}}$ and $M_{d_{joint}}$ are trained for 3 epochs with batch-size of 140 and a learning rate of $5e-7$. For evaluation, the thresholds $T_{dft}$ and $T_{joint}$ are controlled by the fraction of samples routed to $M_{t}$ ($p_{t}\%$).

\noindent
\textbf{Baselines.} 
To the best of our knowledge, we are the first to introduce instance-level routing for multimodal reasoning. Accordingly, we compare against a random routing baseline and two paradigms described in \cref{sec:preliminary}: 
\noindent
\begin{itemize}
\item[$\bullet$]\textcolor{random}{\textbf{Random}}. We use our trained $\mathcal{M}_{d_{dft}}$ or $\mathcal{M}_{d_{joint}}$ to inference and assign instances randomly to draft and target according to $p_t\%$.

\noindent
\item[$\bullet$]\textcolor{P1}{\textbf{Token-probability-based ($Paradigm_1$)}}. On ChartQA, we adopt Qwen2.5-VL-3B as the baseline; on MathVerse and MathVista, we use the GRPO-trained model as the baseline. For each example, we compute $s_1$, sort instances by $s_1$, and perform routing based on this ranking. 

\noindent
\item[$\bullet$]\textcolor{P2}{\textbf{Special-token-based ($Paradigm_2$)}}. On ChartQA, we initialize training from Qwen2.5-VL-3B; on MathVerse and MathVista, we initialize from the GRPO-trained model. For training, each question follows the same sampling budget as our main training setup: we sample $N$ times per question and append  \texttt{[CN]} for correct and \texttt{[UN]} for incorrect, to perform SFT for 3 epochs. At inference, we rank instances by $s_2$, and route accordingly.

\end{itemize}

\begin{table}[t]
    \centering

        \caption{\small Spearman Correlations between rating rank and accuracy show the superiority of our ratings. Area Under the Curve(AUC) for routing demonstrates the overall routing effectiveness.}
        
        \resizebox{1.0\linewidth}{!}{
        \begin{tabular}{c|c|c|c|c|c|c|c|c|c|c|c|c|c|c|c}
            \toprule
            \textbf{Dataset}&\multicolumn{5}{c|}{\textbf{ChartQA}}&\multicolumn{5}{c|}{\textbf{MathVerse}}&\multicolumn{5}{c}{\textbf{MathVista}}\\ \midrule
            \textbf{Paradigm} &\textcolor{random}{\textbf{Random}} &\textcolor{P1}{\textbf{$P_1$}} & \textcolor{P2}{\textbf{$P_2$}} & \textbf{DRL}&\textbf{JRL}&\textcolor{random}{\textbf{Random}} &\textcolor{P1}{\textbf{$P_1$}} & \textcolor{P2}{\textbf{$P_2$}} & \textbf{DRL}&\textbf{JRL}&\textcolor{random}{\textbf{Random}} &\textcolor{P1}{\textbf{$P_1$}} & \textcolor{P2}{\textbf{$P_2$}} & \textbf{DRL}&\textbf{JRL} \\ \midrule

            \textbf{ Spearman of Rank\&Acc\textcolor{red}{$\downarrow$}}&0.00 &-0.34 &-0.25 & \cellcolor{red!10}{\textbf{-0.49}} &  \cellcolor{red!10}{\textbf{-0.50}} & -0.01&-0.01 &-0.11 & \cellcolor{red!10}{\textbf{-0.32}}& \cellcolor{red!10}{\textbf{-0.30}}& 0.00 &0.12 &-0.19 & \cellcolor{red!10}{\textbf{-0.34}} &  \cellcolor{red!10}{\textbf{-0.35}} \\
            \midrule
            
            \textbf{Area \textbf{U}nder \textbf{C}urve(\%)}&78.87&80.74 &73.30 & \cellcolor{red!10}{\textbf{ 81.97}}& \cellcolor{red!10}{\textbf{83.61}} & 56.13&56.72 &56.03 & 55.44& \cellcolor{red!10}{\textbf{56.97}}& 72.09&72.69 &72.19 & \cellcolor{red!10}{\textbf{ 73.29}}& \cellcolor{red!10}{\textbf{73.36}} \\

             \bottomrule
        \end{tabular}
        
        \label{tab:metric} 
        }

\end{table}

\begin{figure*}[t]
  \centering
  \includegraphics[width=1.0\linewidth]{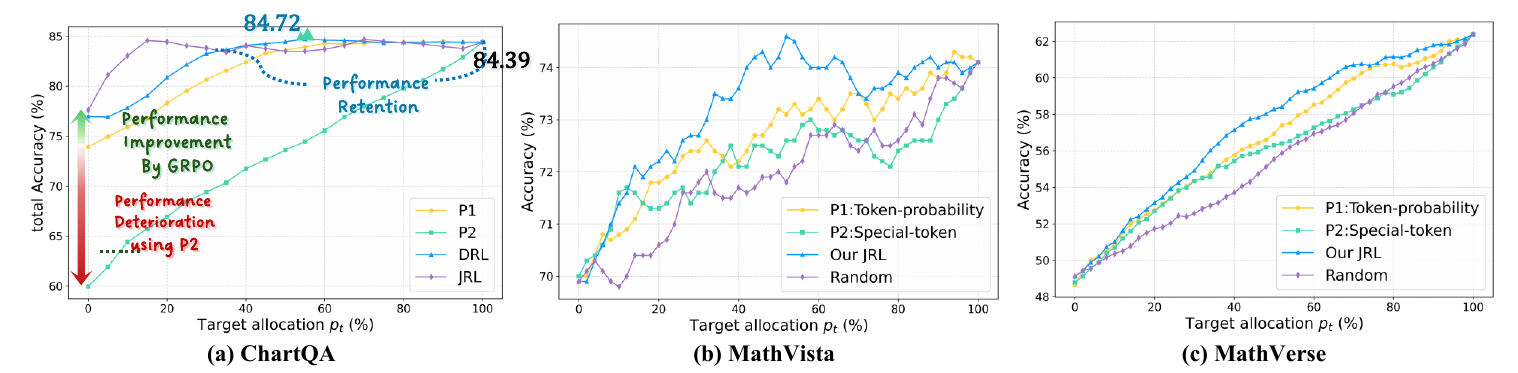}
   \caption{Detailed comparison of our proposed routing with random and 2 previous paradigm on ChartQA, MathVista, and MathVerse.}
   \label{fig:score_exp}
\end{figure*}

\vspace{-0.2cm}
\subsection{Overall Rating and Routing Performance Analysis}
As shown in \cref{tab:metric}, we evaluate the overall rating and routing performance of various paradigms across 3 benchmarks.
\textbf{Overall Rating Performance:} We calculates the Spearman correlation coefficient between the ratings assigned by the draft model and the actual accuracy (ranked from correct to incorrect). A stronger negative correlation indicates that the ratings more accurately reflects its actual performance. Our DRL and JRL consistently achieve higher correlations, proving that our ratings are more closely aligned with true model correctness.
\textbf{Overall Routing Performance:} Area Under the Curve (AUC) is derived from the routing performance curve across various allocation ratios between the draft and target models (See \cref{fig:score_exp}). This reflects the overall effectiveness of the routing process across all possible target allocation $p_t$ levels. We outperform alternative methods in nearly all scenarios, with the notable exception of DRL on MathVista. In this specific case, the DRL performance even falls below the random setting. This occurs because there are instances where the draft model is incapable of solving the task, but the target model also fails; conversely, in tasks where the draft model performs well, the target model exhibits even greater superiority. Because DRL focuses solely on assessing the draft model's capabilities while ignoring the target model's potential, it may incorrectly prioritize samples that would have provided a higher marginal contribution to overall accuracy if assigned differently. We provide a more granular analysis in \cref{sec:failure}.

\subsection{Detailed Evaluation Results on ChartQA}
As shown in \cref{fig:score_exp}(a), we evaluate in-domain DRL-trained $\mathcal{M}_{d_{dft}}$'s and JRL-trained $\mathcal{M}_{d_{joint}}$'s routing capability using scores with Internal Probability Ranking. Compared to \textcolor{P1}{$Paradigm_1$} directly applied to the Qwen2.5-VL-3B model, our DRL-trained $\mathcal{M}_{d_{dft}}$ not only improves model performance but also produces reliable scores, enabling efficient routing. By contrast, \textcolor{P2}{$Paradigm_2$} exhibits a much more severe performance degradation. Notably, DRL can stably maintain, and even enhance, the target model performance from $84.39$ to $84.72$ when $p_t>40$, which demonstrate the feasibility and reliability of our PRP on simple in-domain tasks. JRL can also stably maintain the target model performance when $p_t>15$.

\subsection{Detailed Evaluation Results on Math Domain}

\noindent
We present the detailed results of our Joint Rating Learning with Target Model (JRL) strategy compared with Random, \textcolor{P1}{$Paradigm_1$}, and \textcolor{P2}{$Paradigm_2$} in \cref{fig:score_exp}.
Our JRL consistently outperforms all other strategies across different target allocation ratios, with even larger gains when more queries are routed to the target model. This demonstrates that JRL is a more effective routing mechanism for MLLMs and directly addresses the limitations discussed in ~\cref{subsec:common_limitation}.

\noindent
\textbf{Efficiency Measurements.}
Table~\ref{tab:main_efficiency} summarizes the performance and speedup of our DRL and JRL strategies on the MathVista and MathVerse benchmarks.
For Draft Rating Learning (DRL), we observe that 
the overall performance with DRL-based routing
achieves lossless performance on MathVista with approximately 1.5× acceleration, demonstrating the effectiveness of confidence learning.
With JRL, the routing performance improves substantially once the target model’s capability is incorporated. For example, on MathVerse, $\mathcal{M}_{d_{joint}}$ maintains a competitive 60.4\% accuracy with a 1.42× speedup. On MathVista, JRL performs even better: at a 60\% routing ratio, $\mathcal{M}_{d_{joint}}$ surpasses $\mathcal{M}_{t}$ while achieving a 2.41× speedup.

\begin{table*}[t]
\begin{center}
\caption{Latency and Accuracy evaluations of our DRL and JRL along with baseline methods on MathVista and MathVerse benchmarks.
}
\vspace{-0.5cm}

\resizebox{1\linewidth}{!}{

\begin{tabular}{l|cccc|cccc}
\toprule
\multirow{2}{*}{\textbf{Model}}                                                                               & \multicolumn{4}{c|}{\textbf{MathVista}}                                                                  & \multicolumn{4}{c}{\textbf{MathVerse}}                                                                   \\ \cmidrule{2-9} 
                                                                                                              & \textbf{Draft $(1-p_t)\%$} &  \textbf{Accuracy} & \textbf{Avg. Latency(s)} & \textbf{Speedup} & \textbf{Draft $(1-p_t)\%$}  & \textbf{Accuracy} & \textbf{Avg. Latency(s)} & \textbf{Speedup} \\ \midrule
\textbf{$\mathcal{M}_t$(Skywork-R1V-38B)}                                                                     & -                  &  74.1\%           & 64.8                     & -             & -                  & 62.4\%           & 101.4                     & -                 \\ \midrule

\multirow{4}{*}{\textbf{\begin{tabular}[c]{@{}l@{}}  DRL-trained\\$\mathcal{M}_{d_{dft}}$ (7B)\end{tabular}}} & 100\%                         & 69.9\%           & 10.2                     & -                & 100\%                          & 50.0\%            & 11.8                       & -               \\
                                                                                                               & 20\%                          & \cellcolor{red!10}  75.4\%           & 55.5                       & 1.17x            & 20\%                     & 59.8\%           & 86.4                       & 1.17x            \\
                                                                                                               & 30\%                       & \cellcolor{red!10} 75.3\%           & 53.5                     & 1.21x            & 30\%                         & 58.7\%           & 78.1                       & 1.30x            \\
                                                                                                               & 50\%                         & \cellcolor{red!10} 74.3\%           & 43.6                    & 1.49x            & 40\%                        & 57.2\%           & 69.1                      & 1.47x            \\ \midrule
\multirow{5}{*}{\textbf{\begin{tabular}[c]{@{}l@{}}JRL-trained \\$\mathcal{M}_{d_{joint}}$(7B)\end{tabular}}} & 100\%                           & 70.0\%           & 4.9                      & -                & 100\%                        & 49.9\%           & 10.9                      & -                    \\
                                                                                                               & 30\%                         & \cellcolor{red!10} 74.4\%           & 47.6                     & 1.36x             & 20\%                       & 61.2\%           & 82.1                       & 1.24x            \\
                                                                                                               & 40\%                         & \cellcolor{red!10} 74.1\%           & 40.1                       & 1.62x             & 25\%                      & 60.77\%           & 76.6                       & 1.32x            \\
                                                                                                               & 50\%                          & 74.0\%           & 32.6                      & 1.99x             & 30\%                     & 60.4\%           & 71.5                       & 1.42x            \\
                                                                                                               & \cellcolor{red!10}  60\%                            & \cellcolor{red!10} 74.2\%           & \cellcolor{red!10} 26.9                       & \cellcolor{red!10} 2.41x             & 
                                                                                                                 \cellcolor{red!10} 50\%           & \cellcolor{red!10} 58.5\%           & \cellcolor{red!10}  55.4                       & \cellcolor{red!10}  1.83x            \\
                                                                                                                         \midrule
\textbf{\textcolor{random}{Random}}                                                                                                & 60\%                      & 71.6\%           & 27.5                    & 2.36x            & 50\%                     & 56.4\%           & 56.0                       & 1.81x            \\
\textbf{\textcolor{P1}{Token-probability-based ($Paradigm_1$)}}                                                               & 60\%                         & 72.2\%           & 29.5                       & 2.19x            & 50\%                       & 56.9\%           & 56.3                       & 1.80x            \\
\textbf{\textcolor{P2}{Special-token-based ($Paradigm_2$)}}                                                                   & 60\%                          & 72.1\%           & 26.9                       & \textbf{2.41x}            & 50\%                     & 56.3\%           & 55.6                      & 1.82x            \\

\rowcolor{red!10}\textbf{Ours}                 & 60\%                    & \textbf{74.2\%}           & \textbf{26.9 }                    & \textbf{2.41x}             &  50\%             &  \textbf{58.5\%}           &   \textbf{55.4}              & \textbf{1.83x }           \\
                                                                                                               \bottomrule
\end{tabular}
}

\label{tab:main_efficiency}
\end{center}
\vspace{-0.3cm}
\end{table*}

\subsection{Ablation Studies}
\label{sec:ablation}

\begin{figure*}[t]
  \centering
  \includegraphics[width=1\linewidth]{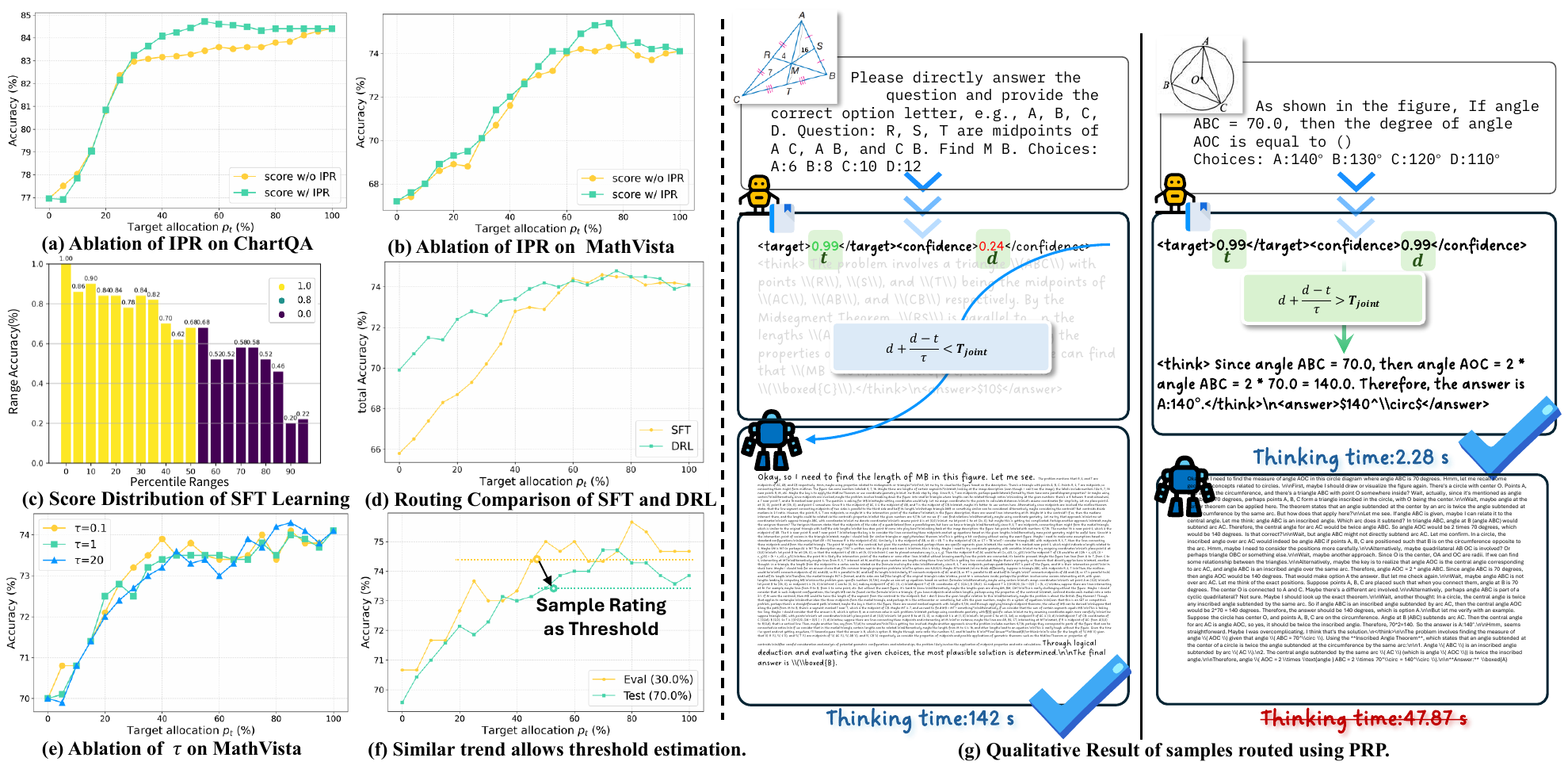}
   \caption{Ablation Study: (a)(b) Ablation study of Internal Probability Ranking (IPR) on ChartQA and MathVista.(c) Score Distribution of SFT Rating Learning. (d) Routing Comparison of SFT and DRL. (e). $\mathcal{M}_{d_{joint}}$'s ablation of different $\tau$ for the scaling in signal $s=d+\frac{d-t}{\tau}$ on MathVista. (f) Similar accuracy trends across evaluation and test sets allows threshold estimation. (g) Qualitative Result of samples routed using PRP.}

   \label{fig:ablations}
\end{figure*}

\noindent
\textbf{Internal Probability Ranking (IPR)}. 
We validate the effectiveness of IPR on ChartQA and MathVista. Specifically, we compare (i) routing based solely on raw confidence ratings, and (ii) routing with additional Internal Probability Ranking within each score range. As shown in \cref{fig:ablations}(a)(b), the latter consistently achieves better routing quality due to the more orderly score distribution within each bin. This demonstrates that incorporating internal probability provides finer-grained confidence calibration for the draft model, leading to improved performance.

\noindent
\textbf{RL-based rating training vs SFT-based rating training}. 
As a comparison, we use the accuracy within the group as the label and employ Supervised Finetuning for rating training. Compared to our rich scoring scenarios, the SFT model produces polarized scores (only 0/1) during inference (see \cref{fig:ablations}(c)), resulting in a weaker correlation with actual performance than our RL-based method. Meanwhile, the draft accuracy on MathVista drops by 4.2\%(see \cref{fig:ablations}(d)).

\noindent
\textbf{Different scale $\tau$ for $\mathcal{M}_{d_{joint}}$}.
As discussed in \cref{sec:routing}, when employing $\mathcal{M}_{d_{joint}}$ for routing, the scaling coefficient $\tau$ controls the importance of the gap between $\mathcal{M}_{d_{joint}}$ and $\mathcal{M}_{t}$ during ranking. As shwon in \cref{fig:ablations}(e),When $\tau$ is large (blue curve), routing focuses more on $\mathcal{M}_{d_{joint}}$’s intrinsic capability, thus preferentially assigning instances where $\mathcal{M}_{d_{joint}}$ excels to the $\mathcal{M}_{d_{joint}}$; under this setting, $p_t\in[80,100]$ exhibits strong performance preservation. Conversely, when $\tau$ is small (yellow curve), routing emphasizes the discrepancy between $\mathcal{M}_{d_{joint}}$ and $\mathcal{M}_{t}$, preferentially preserving instances where $\mathcal{M}_{t}$ excels to $\mathcal{M}_{t}$; as a result, we achieve faster performance gains in $p_t \in [0, 50]$.

\noindent
\textbf{Qualitative Result}.
In \cref{fig:ablations}(g), PRP routing proceeds once module $\mathcal{M}_{d_{dft}}$ produces a self-score $d$ and an assessment 
$t$ of 
$\mathcal{M}_t$. 
More results are in \cref{sec:more_quali}.

\vspace{-0.1cm}

\section{Related Work}
\label{sec:relatedwork}

\subsection{Reinforcement Learning for Reasoning Models} 
Reinforcement learning has been widely adopted to enhance the reasoning capabilities of LLMs. Methods such as PPO\cite{ppo} and DPO\cite{dpo} are commonly used in post-training by maximizing reward functions. 
GRPO\cite{deepseekr1,deepseekmath} was proposed to improve reasoning by optimizing group advantages using reward differences across multiple completions, while subsequent works\cite{drgrpo,yu2025dapo,meng2025mmeureka,liu2025cpgd} explore generalized optimization strategies. 
GRPO-like approaches have been extended to multimodal domains, where customized reward functions are designed for various vision-language tasks\cite{chen2025grpocare,zhang2025r1vl,shen2025vlmr1,du2025mmprm,chen2025g1bootstrappingperceptionreasoning,liu2025visual,liu2025visualagenticreinforcementfinetuning}.
We leverage GRPO to preserve draft model's performance while equipping the model with proactive rating ability, thereby laying the foundation for flexible routing.

\subsection{Efficient Reasoning}
LLM reasoning often requires long chains of thought, leading to increased latency\cite{qu2025survey}. 
A common solution is to pair a small draft model with a larger target model. 
Speculative decoding\cite{leviathan2023fast_sd,chen2023accelerating_sd,cai2024medusa,li2024eagle,miao2024specinfer,elhoushi2024layerskip} is one representative approach, but applying it to reasoning models introduces token-distribution mismatch across intermediate steps. 
Progress reward models (PRMs)\cite{liao2025rsd,wang2025SpecSearch} attempt to assign reasoning steps to draft or target models, but they are difficult to train for MLLMs and incur additional cost. 
Router-based methods\cite{chuang2024self-ref,feng2024graphrouter,fu2025r2r,ong2024routellm,barrak2025cargoframeworkconfidenceawarerouting} rely on evaluating the model’s confidence\cite{geng-etal-2024-survey}, and offer another direction by selecting models according to query difficulty or budget. 
For example, Self-REF\cite{chuang2024self-ref} uses SFT to train a confidence token, while others\cite{chuang2025confident} benchmark uncertainty estimation for reasoning LLMs. 
However, their effectiveness for MLLMs remains unclear; they typically rely on full responses, preventing early routing, and ignore the target model’s solvability. 
In this work, we propose to estimate  confidence \emph{before} reasoning, enabling MLLMs to decide when and whom to ask.

\vspace{-0.1cm}
\section{Conclusion}
\label{sec:conclusion}
\vspace{-0.1cm}
We present a RL-based Proactive Routing Paradigm for efficient MLLM reasoning. By introducing Draft Rating Learning and Joint Rating Learning, we enables models to estimate both their own and the target model’s solvability of the input before generation, making proactive and capability-aware routing possible. Building on robust rating model, our three-stage PRP enables fine-grained routing that prioritizes samples for which the target model is expected to excel, rather than simply forwarding the most difficult cases, yielding a favorable balance between performance and efficiency across the draft and target models.

\section*{Acknowledgements}
This work is supported by National Key Research and Development Program of China 
(2023YFC3321600), the Early Career Scheme (No.CityU 21219323) and the General Research Fund (No.CityU 11220324) of the University Grants Committee (UGC), the NSFC Young Scientists Fund (No.9240127), and the Donation for Research Projects (No.9229164 and No.9229216).

\clearpage

\bibliographystyle{splncs04}
\bibliography{main}

@String(CVPR  = {IEEE Conf. Comput. Vis. Pattern Recog.})

@String(ICCV  = {Int. Conf. Comput. Vis.})

@String(ICLR  = {Int. Conf. Learn. Represent.})

@String(CVPR  = {CVPR})

@String(ICCV  = {ICCV})

@String(ICLR  = {ICLR})

@article{dpo,
  title={Direct preference optimization: Your language model is secretly a reward model},
  author={Rafailov, Rafael and Sharma, Archit and Mitchell, Eric and Manning, Christopher D and Ermon, Stefano and Finn, Chelsea},
  journal={Advances in Neural Information Processing Systems},
  volume={36},
  pages={53728--53741},
  year={2023}
}

@article{ppo,
  title={Proximal policy optimization algorithms},
  author={Schulman, John and Wolski, Filip and Dhariwal, Prafulla and Radford, Alec and Klimov, Oleg},
  journal={arXiv preprint arXiv:1707.06347},
  year={2017}
}

@article{deepseekmath,
  title={Deepseekmath: Pushing the limits of mathematical reasoning in open language models},
  author={Shao, Zhihong and Wang, Peiyi and Zhu, Qihao and Xu, Runxin and Song, Junxiao and Bi, Xiao and Zhang, Haowei and Zhang, Mingchuan and Li, YK and Wu, Y and others},
  journal={arXiv preprint arXiv:2402.03300},
  year={2024}
}

@article{deepseekr1,
  title={Deepseek-r1: Incentivizing reasoning capability in llms via reinforcement learning},
  author={Guo, Daya and Yang, Dejian and Zhang, Haowei and Song, Junxiao and Zhang, Ruoyu and Xu, Runxin and Zhu, Qihao and Ma, Shirong and Wang, Peiyi and Bi, Xiao and others},
  journal={arXiv preprint arXiv:2501.12948},
  year={2025}
}

@article{meng2025mmeureka,
      title={MM-Eureka: Exploring the Frontiers of Multimodal Reasoning with Rule-based Reinforcement Learning},
      author={Fanqing Meng and Lingxiao Du and Zongkai Liu and Zhixiang Zhou and Quanfeng Lu and Daocheng Fu and Tiancheng Han and Botian Shi and Wenhai Wang and Junjun He and Kaipeng Zhang and Ping Luo and Yu Qiao and Qiaosheng Zhang and Wenqi Shao},
      year={2025},
      journal={arXiv preprint arXiv:2503.07365},
}

@article{du2025mmprm,
      title={MM-PRM: Enhancing Multimodal Mathematical Reasoning with Scalable Step-Level Supervision},
      author={Lingxiao Du and Fanqing Meng and Zongkai Liu and Zhixiang Zhou and Ping Luo and Qiaosheng Zhang and Wenqi Shao},
      year={2025},
      journal={arXiv preprint arXiv:2505.13427},
}

@article{liu2025cpgd,
      title={CPGD: Toward Stable Rule-based Reinforcement Learning for Language Models},
      author={Zongkai Liu and Fanqing Meng and Lingxiao Du and Zhixiang Zhou and Chao Yu and Wenqi Shao and Qiaosheng Zhang},
      year={2025},
      journal={arXiv preprint arXiv:2505.12504},
}

@article{shen2025vlmr1,
  title={Vlm-r1: A stable and generalizable r1-style large vision-language model},
  author={Shen, Haozhan and Liu, Peng and Li, Jingcheng and Fang, Chunxin and Ma, Yibo and Liao, Jiajia and Shen, Qiaoli and Zhang, Zilun and Zhao, Kangjia and Zhang, Qianqian and others},
  journal={arXiv preprint arXiv:2504.07615},
  year={2025}
}

@article{yu2025dapo,
  title={Dapo: An open-source llm reinforcement learning system at scale},
  author={Yu, Qiying and Zhang, Zheng and Zhu, Ruofei and Yuan, Yufeng and Zuo, Xiaochen and Yue, Yu and Dai, Weinan and Fan, Tiantian and Liu, Gaohong and Liu, Lingjun and others},
  journal={arXiv preprint arXiv:2503.14476},
  year={2025}
}

@article{drgrpo,
  title={Understanding r1-zero-like training: A critical perspective},
  author={Liu, Zichen and Chen, Changyu and Li, Wenjun and Qi, Penghui and Pang, Tianyu and Du, Chao and Lee, Wee Sun and Lin, Min},
  journal={arXiv preprint arXiv:2503.20783},
  year={2025}
}

@article{chen2025grpocare,
  title={GRPO-CARE: Consistency-Aware Reinforcement Learning for Multimodal Reasoning},
  author={Chen, Yi and Ge, Yuying and Wang, Rui and Ge, Yixiao and Cheng, Junhao and Shan, Ying and Liu, Xihui},
  journal={arXiv preprint arXiv:2506.16141},
  year={2025}
}

@article{zhang2025r1vl,
  title={R1-vl: Learning to reason with multimodal large language models via step-wise group relative policy optimization},
  author={Zhang, Jingyi and Huang, Jiaxing and Yao, Huanjin and Liu, Shunyu and Zhang, Xikun and Lu, Shijian and Tao, Dacheng},
  journal={arXiv preprint arXiv:2503.12937},
  year={2025}
}

@article{liu2025visual,
  title={Visual-rft: Visual reinforcement fine-tuning},
  author={Liu, Ziyu and Sun, Zeyi and Zang, Yuhang and Dong, Xiaoyi and Cao, Yuhang and Duan, Haodong and Lin, Dahua and Wang, Jiaqi},
  journal={arXiv preprint arXiv:2503.01785},
  year={2025}
}

@inproceedings{leviathan2023fast_sd,
  title={Fast inference from transformers via speculative decoding},
  author={Leviathan, Yaniv and Kalman, Matan and Matias, Yossi},
  booktitle={International Conference on Machine Learning},
  pages={19274--19286},
  year={2023},
  organization={PMLR}
}

@article{chen2023accelerating_sd,
  title={Accelerating large language model decoding with speculative sampling},
  author={Chen, Charlie and Borgeaud, Sebastian and Irving, Geoffrey and Lespiau, Jean-Baptiste and Sifre, Laurent and Jumper, John},
  journal={arXiv preprint arXiv:2302.01318},
  year={2023}
}

@inproceedings{miao2024specinfer,
  title={Specinfer: Accelerating large language model serving with tree-based speculative inference and verification},
  author={Miao, Xupeng and Oliaro, Gabriele and Zhang, Zhihao and Cheng, Xinhao and Wang, Zeyu and Zhang, Zhengxin and Wong, Rae Ying Yee and Zhu, Alan and Yang, Lijie and Shi, Xiaoxiang and others},
  booktitle={Proceedings of the 29th ACM International Conference on Architectural Support for Programming Languages and Operating Systems, Volume 3},
  pages={932--949},
  year={2024}
}

@article{elhoushi2024layerskip,
  title={LayerSkip: Enabling early exit inference and self-speculative decoding},
  author={Elhoushi, Mostafa and Shrivastava, Akshat and Liskovich, Diana and Hosmer, Basil and Wasti, Bram and Lai, Liangzhen and Mahmoud, Anas and Acun, Bilge and Agarwal, Saurabh and Roman, Ahmed and others},
  journal={arXiv preprint arXiv:2404.16710},
  year={2024}
}

@article{cai2024medusa,
  title={Medusa: Simple llm inference acceleration framework with multiple decoding heads},
  author={Cai, Tianle and Li, Yuhong and Geng, Zhengyang and Peng, Hongwu and Lee, Jason D and Chen, Deming and Dao, Tri},
  journal={arXiv preprint arXiv:2401.10774},
  year={2024}
}

@article{li2024eagle,
  title={Eagle: Speculative sampling requires rethinking feature uncertainty},
  author={Li, Yuhui and Wei, Fangyun and Zhang, Chao and Zhang, Hongyang},
  journal={arXiv preprint arXiv:2401.15077},
  year={2024}
}

@article{liao2025rsd,
  title={Reward-guided speculative decoding for efficient llm reasoning},
  author={Liao, Baohao and Xu, Yuhui and Dong, Hanze and Li, Junnan and Monz, Christof and Savarese, Silvio and Sahoo, Doyen and Xiong, Caiming},
  journal={arXiv preprint arXiv:2501.19324},
  year={2025}
}

@article{wang2025SpecSearch,
  title={Accelerating Large Language Model Reasoning via Speculative Search},
  author={Wang, Zhihai and Wang, Jie and Pan, Jilai and Xia, Xilin and Zhen, Huiling and Yuan, Mingxuan and Hao, Jianye and Wu, Feng},
  journal={arXiv preprint arXiv:2505.02865},
  year={2025}
}

@misc{chen2025g1bootstrappingperceptionreasoning,
      title={G1: Bootstrapping Perception and Reasoning Abilities of Vision-Language Model via Reinforcement Learning}, 
      author={Liang Chen and Hongcheng Gao and Tianyu Liu and Zhiqi Huang and Flood Sung and Xinyu Zhou and Yuxin Wu and Baobao Chang},
      year={2025},
      eprint={2505.13426},
      archivePrefix={arXiv},
      primaryClass={cs.CV},
      url={https://arxiv.org/abs/2505.13426}, 
}

@misc{liu2025visualagenticreinforcementfinetuning,
      title={Visual Agentic Reinforcement Fine-Tuning}, 
      author={Ziyu Liu and Yuhang Zang and Yushan Zou and Zijian Liang and Xiaoyi Dong and Yuhang Cao and Haodong Duan and Dahua Lin and Jiaqi Wang},
      year={2025},
      eprint={2505.14246},
      archivePrefix={arXiv},
      primaryClass={cs.CV},
      url={https://arxiv.org/abs/2505.14246}, 
}

@inproceedings{lu2024mathvista,
  title={MathVista: Evaluating Mathematical Reasoning of Foundation Models in Visual Contexts},
  author={Lu, Pan and Bansal, Hritik and Xia, Tony and Liu, Jiacheng and Li, Chunyuan and Hajishirzi, Hannaneh and Cheng, Hao and Chang, Kai-Wei and Galley, Michel and Gao, Jianfeng},
  booktitle={International Conference on Learning Representations (ICLR)},
  year={2024}
}

@article{zhang2024mathverse,
  title={MathVerse: Does Your Multi-modal LLM Truly See the Diagrams in Visual Math Problems?},
  author={Zhang, Renrui and Jiang, Dongzhi and Zhang, Yichi and Lin, Haokun and Guo, Ziyu and Qiu, Pengshuo and Zhou, Aojun and Lu, Pan and Chang, Kai-Wei and Gao, Peng and others},
  journal={arXiv preprint arXiv:2403.14624},
  year={2024}
}

@inproceedings{masry2022chartqa,
  title={Chartqa: A benchmark for question answering about charts with visual and logical reasoning},
  author={Masry, Ahmed and Do, Xuan Long and Tan, Jia Qing and Joty, Shafiq and Hoque, Enamul},
  booktitle={Findings of the association for computational linguistics: ACL 2022},
  pages={2263--2279},
  year={2022}
}

@inproceedings{mathew2021docvqa,
  title={Docvqa: A dataset for vqa on document images},
  author={Mathew, Minesh and Karatzas, Dimosthenis and Jawahar, CV},
  booktitle={Proceedings of the IEEE/CVF winter conference on applications of computer vision},
  pages={2200--2209},
  year={2021}
}

@article{bai2025qwen2.5vl,
  title={Qwen2. 5-vl technical report},
  author={Bai, Shuai and Chen, Keqin and Liu, Xuejing and Wang, Jialin and Ge, Wenbin and Song, Sibo and Dang, Kai and Wang, Peng and Wang, Shijie and Tang, Jun and others},
  journal={arXiv preprint arXiv:2502.13923},
  year={2025}
}

@article{lu2025ovis2.5,
  title={Ovis2. 5 technical report},
  author={Lu, Shiyin and Li, Yang and Xia, Yu and Hu, Yuwei and Zhao, Shanshan and Ma, Yanqing and Wei, Zhichao and Li, Yinglun and Duan, Lunhao and Zhao, Jianshan and others},
  journal={arXiv preprint arXiv:2508.11737},
  year={2025}
}

@misc{shen2025skyworkr1v3technicalreport,
      title={Skywork-R1V3 Technical Report}, 
      author={Wei Shen and Jiangbo Pei and Yi Peng and Xuchen Song and Yang Liu and Jian Peng and Haofeng Sun and Yunzhuo Hao and Peiyu Wang and Jianhao Zhang and Yahui Zhou},
      year={2025},
      eprint={2507.06167},
      archivePrefix={arXiv},
      primaryClass={cs.CL},
      url={https://arxiv.org/abs/2507.06167}, 
}

@article{chuang2024self-ref,
  title={Learning to route llms with confidence tokens},
  author={Chuang, Yu-Neng and Sarma, Prathusha Kameswara and Gopalan, Parikshit and Boccio, John and Bolouki, Sara and Hu, Xia and Zhou, Helen},
  journal={arXiv preprint arXiv:2410.13284},
  year={2024}
}

@article{chuang2025confident,
  title={Confident or seek stronger: Exploring uncertainty-based on-device llm routing from benchmarking to generalization},
  author={Chuang, Yu-Neng and Yu, Leisheng and Wang, Guanchu and Zhang, Lizhe and Liu, Zirui and Cai, Xuanting and Sui, Yang and Braverman, Vladimir and Hu, Xia},
  journal={arXiv preprint arXiv:2502.04428},
  year={2025}
}

@article{mahaut2024factual,
  title={Factual confidence of llms: on reliability and robustness of current estimators},
  author={Mahaut, Mat{\'e}o and Aina, Laura and Czarnowska, Paula and Hardalov, Momchil and M{\"u}ller, Thomas and M{\`a}rquez, Llu{\'\i}s},
  journal={arXiv preprint arXiv:2406.13415},
  year={2024}
}

@article{feng2024graphrouter,
  title={Graphrouter: A graph-based router for llm selections},
  author={Feng, Tao and Shen, Yanzhen and You, Jiaxuan},
  journal={arXiv preprint arXiv:2410.03834},
  year={2024}
}

@article{qu2025survey,
  title={A survey of efficient reasoning for large reasoning models: Language, multimodality, and beyond},
  author={Qu, Xiaoye and Li, Yafu and Su, Zhaochen and Sun, Weigao and Yan, Jianhao and Liu, Dongrui and Cui, Ganqu and Liu, Daizong and Liang, Shuxian and He, Junxian and others},
  journal={arXiv preprint arXiv:2503.21614},
  year={2025}
}

@article{fu2025r2r,
  title={R2R: Efficiently Navigating Divergent Reasoning Paths with Small-Large Model Token Routing},
  author={Fu, Tianyu and Ge, Yi and You, Yichen and Liu, Enshu and Yuan, Zhihang and Dai, Guohao and Yan, Shengen and Yang, Huazhong and Wang, Yu},
  journal={arXiv preprint arXiv:2505.21600},
  year={2025}
}

@article{ong2024routellm,
  title={Routellm: Learning to route llms with preference data},
  author={Ong, Isaac and Almahairi, Amjad and Wu, Vincent and Chiang, Wei-Lin and Wu, Tianhao and Gonzalez, Joseph E and Kadous, M Waleed and Stoica, Ion},
  journal={arXiv preprint arXiv:2406.18665},
  year={2024}
}

@article{liu2023visual,
  title={Visual instruction tuning},
  author={Liu, Haotian and Li, Chunyuan and Wu, Qingyang and Lee, Yong Jae},
  journal={Advances in neural information processing systems},
  volume={36},
  pages={34892--34916},
  year={2023}
}

@article{zhu2025internvl3,
  title={Internvl3: Exploring advanced training and test-time recipes for open-source multimodal models},
  author={Zhu, Jinguo and Wang, Weiyun and Chen, Zhe and Liu, Zhaoyang and Ye, Shenglong and Gu, Lixin and Tian, Hao and Duan, Yuchen and Su, Weijie and Shao, Jie and others},
  journal={arXiv preprint arXiv:2504.10479},
  year={2025}
}

@article{chen2024expanding,
  title={Expanding Performance Boundaries of Open-Source Multimodal Models with Model, Data, and Test-Time Scaling},
  author={Chen, Zhe and Wang, Weiyun and Cao, Yue and Liu, Yangzhou and Gao, Zhangwei and Cui, Erfei and Zhu, Jinguo and Ye, Shenglong and Tian, Hao and Liu, Zhaoyang and others},
  journal={arXiv preprint arXiv:2412.05271},
  year={2024}
}

@article{comanici2025gemini,
  title={Gemini 2.5: Pushing the frontier with advanced reasoning, multimodality, long context, and next generation agentic capabilities},
  author={Comanici, Gheorghe and Bieber, Eric and Schaekermann, Mike and Pasupat, Ice and Sachdeva, Noveen and Dhillon, Inderjit and Blistein, Marcel and Ram, Ori and Zhang, Dan and Rosen, Evan and others},
  journal={arXiv preprint arXiv:2507.06261},
  year={2025}
}

@article{hurst2024gpt,
  title={Gpt-4o system card},
  author={Hurst, Aaron and Lerer, Adam and Goucher, Adam P and Perelman, Adam and Ramesh, Aditya and Clark, Aidan and Ostrow, AJ and Welihinda, Akila and Hayes, Alan and Radford, Alec and others},
  journal={arXiv preprint arXiv:2410.21276},
  year={2024}
}

@article{peng2025lmm,
  title={Lmm-r1: Empowering 3b lmms with strong reasoning abilities through two-stage rule-based rl},
  author={Peng, Yingzhe and Zhang, Gongrui and Zhang, Miaosen and You, Zhiyuan and Liu, Jie and Zhu, Qipeng and Yang, Kai and Xu, Xingzhong and Geng, Xin and Yang, Xu},
  journal={arXiv preprint arXiv:2503.07536},
  year={2025}
}

@misc{chen2025r1v,
  author       = {Chen, Liang and Li, Lei and Zhao, Haozhe and Song, Yifan and Vinci},
  title        = {R1-V: Reinforcing Super Generalization Ability in Vision-Language Models with Less Than \$3},
  howpublished = {\url{https://github.com/Deep-Agent/R1-V}},
  note         = {Accessed: 2025-02-02},
  year         = {2025}
}

@misc{openr1,
    title = {Open R1: A fully open reproduction of DeepSeek-R1},
    url = {https://github.com/huggingface/open-r1},
    author = {{Hugging Face}},
    month = {January},
    year = {2025}
}

@misc{wang2025vlrethinkerincentivizingselfreflectionvisionlanguage,
      title={VL-Rethinker: Incentivizing Self-Reflection of Vision-Language Models with Reinforcement Learning}, 
      author={Haozhe Wang and Chao Qu and Zuming Huang and Wei Chu and Fangzhen Lin and Wenhu Chen},
      year={2025},
      eprint={2504.08837},
      archivePrefix={arXiv},
      primaryClass={cs.LG},
      url={https://arxiv.org/abs/2504.08837}, 
}

@inproceedings{m3cot,
    title = "M$^3$CoT: A Novel Benchmark for Multi-Domain Multi-step Multi-modal Chain-of-Thought",
    author = "Chen, Qiguang  and
      Qin, Libo  and
      Zhang, Jin  and
      Chen, Zhi  and
      Xu, Xiao  and
      Che, Wanxiang",
    booktitle = "Proc. of ACL",
    year = "2024",
}

@inproceedings{geng-etal-2024-survey,
    title = "A Survey of Confidence Estimation and Calibration in Large Language Models",
    author = "Geng, Jiahui  and
      Cai, Fengyu  and
      Wang, Yuxia  and
      Koeppl, Heinz  and
      Nakov, Preslav  and
      Gurevych, Iryna",
    editor = "Duh, Kevin  and
      Gomez, Helena  and
      Bethard, Steven",
    booktitle = "Proceedings of the 2024 Conference of the North American Chapter of the Association for Computational Linguistics: Human Language Technologies (Volume 1: Long Papers)",
    month = jun,
    year = "2024",
    address = "Mexico City, Mexico",
    publisher = "Association for Computational Linguistics",
    url = "https://aclanthology.org/2024.naacl-long.366/",
    doi = "10.18653/v1/2024.naacl-long.366",
    pages = "6577--6595"
}

@misc{barrak2025cargoframeworkconfidenceawarerouting,
      title={CARGO: A Framework for Confidence-Aware Routing of Large Language Models}, 
      author={Amine Barrak and Yosr Fourati and Michael Olchawa and Emna Ksontini and Khalil Zoghlami},
      year={2025},
      eprint={2509.14899},
      archivePrefix={arXiv},
      primaryClass={cs.SE},
      url={https://arxiv.org/abs/2509.14899}, 
}

@INPROCEEDINGS{11446010,
  author={Zhou, Yinan and Chen, Yuxin and Lin, Haokun and Wu, Yichen and Yang, Shuyu and Qi, Zhongang and Ma, Chen and Zhu, Li},
  booktitle={2025 IEEE/CVF International Conference on Computer Vision (ICCV)}, 
  title={DOGR: Towards Versatile Visual Document Grounding and Referring}, 
  year={2025},
  volume={},
  number={},
  pages={3596-3606},
  doi={10.1109/ICCV51701.2025.00343}}

@ARTICLE{11125949,
  author={Zhou, Yinan and Wang, Yaxiong and Lin, Haokun and Ma, Chen and Zhu, Li and Zheng, Zhedong},
  journal={IEEE Transactions on Multimedia}, 
  title={Scale Up Composed Image Retrieval Learning via Modification Text Generation}, 
  year={2025},
  volume={27},
  number={},
  pages={7510-7521},
  doi={10.1109/TMM.2025.3599088}}

@inproceedings{10.1145/3581783.3611709,
author = {Yang, Shuyu and Zhou, Yinan and Zheng, Zhedong and Wang, Yaxiong and Zhu, Li and Wu, Yujiao},
title = {Towards Unified Text-based Person Retrieval: A Large-scale Multi-Attribute and Language Search Benchmark},
year = {2023},
isbn = {9798400701085},
publisher = {Association for Computing Machinery},
address = {New York, NY, USA},
url = {https://doi.org/10.1145/3581783.3611709},
doi = {10.1145/3581783.3611709},
booktitle = {Proceedings of the 31st ACM International Conference on Multimedia},
pages = {4492–4501},
numpages = {10},
location = {Ottawa ON, Canada},
series = {MM '23}
}

@misc{yang2025detailfusiondualbranchframeworkenhancement,
      title={DetailFusion: A Dual-branch Framework with Detail Enhancement for Composed Image Retrieval}, 
      author={Yuxin Yang and Yinan Zhou and Yuxin Chen and Ziqi Zhang and Zongyang Ma and Chunfeng Yuan and Bing Li and Lin Song and Jun Gao and Peng Li and Weiming Hu},
      year={2025},
      eprint={2505.17796},
      archivePrefix={arXiv},
      primaryClass={cs.CV},
      url={https://arxiv.org/abs/2505.17796}, 
}

@InProceedings{Yang_2026_CVPR,
    author    = {Yang, Yuxin and Zhou, Yinan and Chen, Yuxin and Zhang, Ziqi and Ma, Zongyang and Yuan, Chunfeng and Li, Bing and Gao, Jun and Hu, Weiming},
    title     = {Beyond Semantic Search: Towards Referential Anchoring in Composed Image Retrieval},
    booktitle = {Proceedings of the IEEE/CVF Conference on Computer Vision and Pattern Recognition (CVPR)},
    month     = {June},
    year      = {2026},
    pages     = {31155-31165}
}

\clearpage
\appendix
\setcounter{page}{1}

\section{More Experiments}
\subsection{Detail Rating distributions}

\begin{figure*}[t]
  \centering

  \includegraphics[width=1\linewidth]{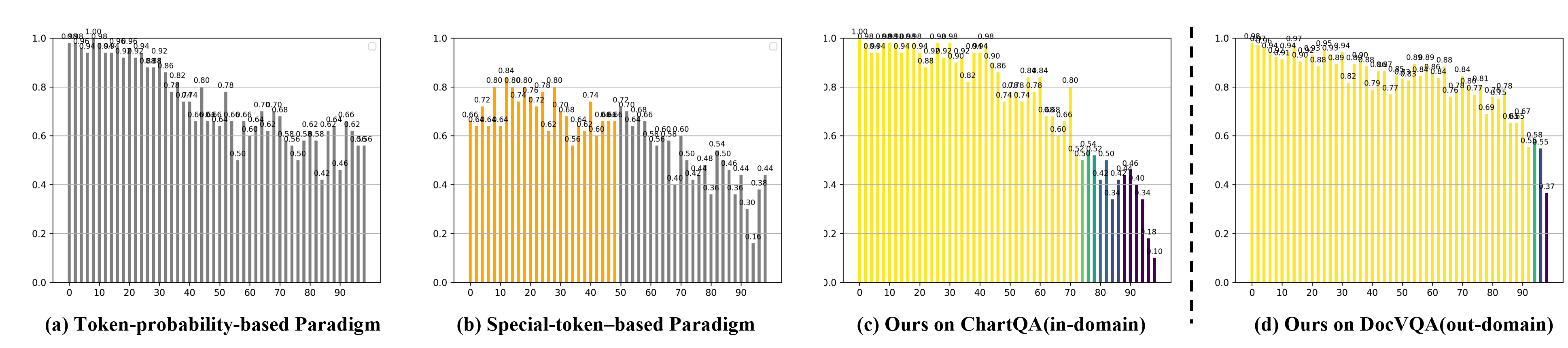}
    \vspace{-0.6cm}

   \caption{Score distribution comparison on ChartQA and ours out-domain disribution on DocVQA.}
    \vspace{-0.4cm}

   \label{fig:chartscore}
\end{figure*}

\noindent
\textbf{ChartQA.} As shown in \cref{fig:chartscore}, on ChartQA both $Paradigm_1$ (a) and $Paradigm_2$ (b) exhibit discernible score-distribution trends. However, the distribution of $Paradigm_2$ in (b) comes at the cost of the draft model’s overall performance. In contrast, our paradigm (c) produces a more pronounced separation in score distributions than either baseline on ChartQA. Furthermore, we evaluate out-of-domain score distributions on DocVQA\cite{mathew2021docvqa} in (d): despite being trained exclusively on chart data, the scoring model maintains a clear separation on document QA task, demonstrating strong generalization of the draft model’s scoring capability.

\begin{figure*}[t]
  \centering

  \includegraphics[width=1\linewidth]{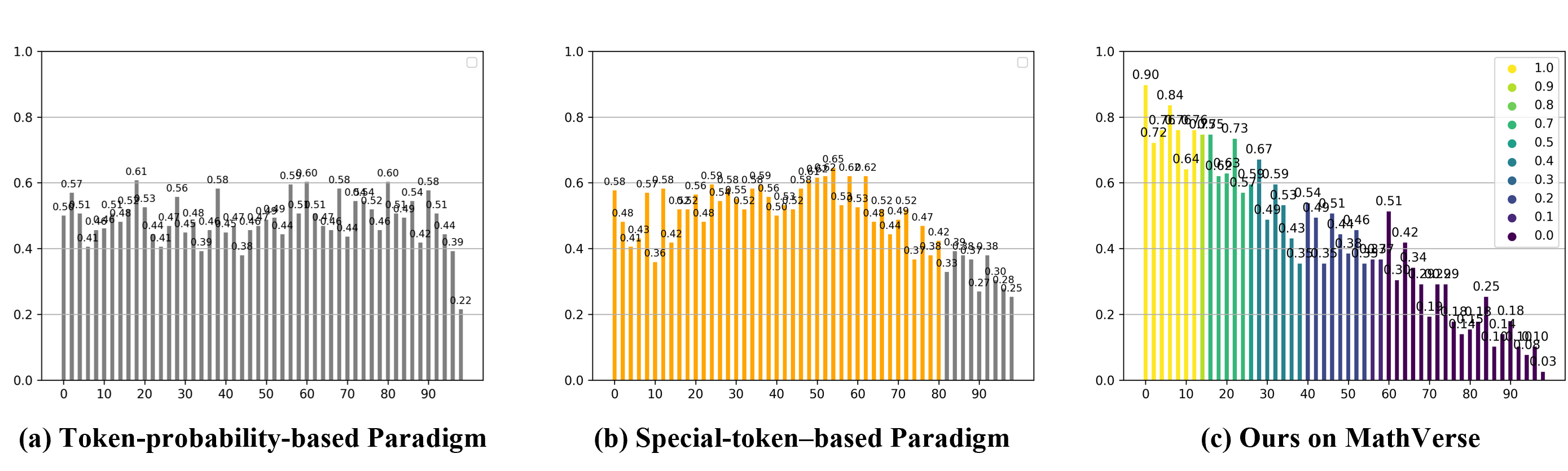}
    \vspace{-0.7cm}

   \caption{Score distribution comparison on MathVerse.}
    \vspace{-0.4cm}

   \label{fig:mathversescore}
\end{figure*}

\noindent
\textbf{MathVerse.} As shown in \cref{fig:mathversescore} on MathVerse, the score distribution under $Paradigm_1$  (a) is notably diffuse. Under $Paradigm_2$ (b), the portion predicted as \texttt{[UN]} exhibits substantially lower accuracy than the portion predicted as  \texttt{[CN]}, yet the overall separability remains limited. In contrast, ours (c) yields a much clearer separation: the scores exhibit a pronounced stepwise trend, forming a clean gradient from 90\% accuracy down to 0.03\% accuracy. This pattern highlights the superior scoring effectiveness of our draft model.

\begin{figure*}[t]
  \centering

  \includegraphics[width=1\linewidth]{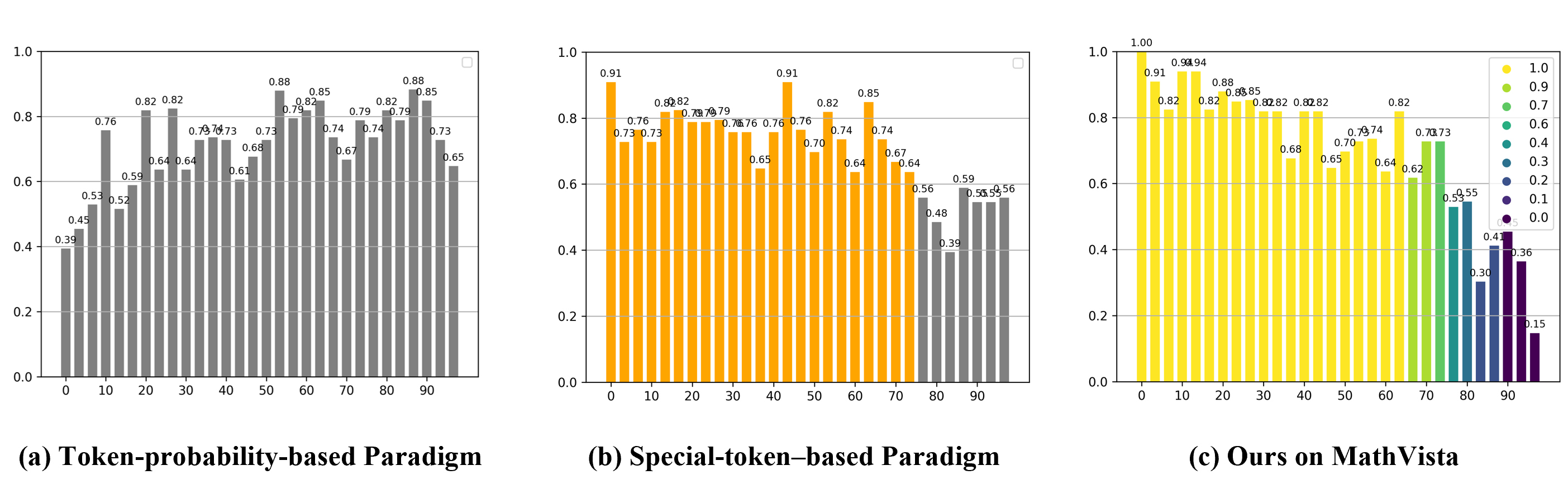}
    \vspace{-0.6cm}

   \caption{Score distribution comparison on MathVista.}
    \vspace{-0.3cm}

   \label{fig:mathvistascore}
\end{figure*}

\noindent
\textbf{MathVista.} As shown in \cref{fig:mathvistascore} on MathVista, $Paradigm_1$ score distribution exhibits the opposite trend: overly long outputs tend to weaken visual grounding and push the model to rely primarily on text tokens, such that examples with larger signal paradoxically achieve lower accuracy. In contrast, $Paradigm_2$ distribution in (b) resembles that on MathVerse, retaining only limited separability. Compared to these two paradigms, our distribution in (c) remains more discriminative, displaying a clear monotonic decline in accuracy from 100\% down to 15\%.

\begin{figure*}[t]
  \centering

  \includegraphics[width=1\linewidth]{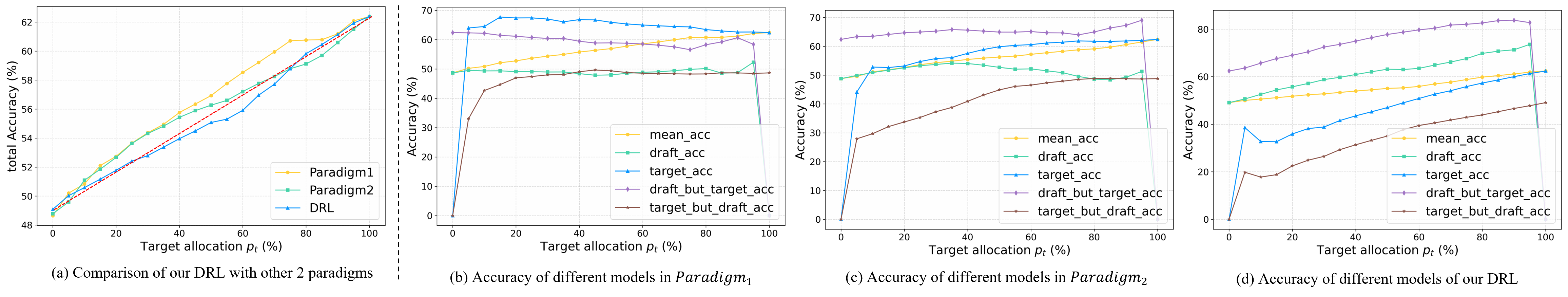}
    \vspace{-0.6cm}

   \caption{Discussion of DRL-based Routing Failure on MathVerse.}
    \vspace{-0.3cm}
   
   \label{fig:drlfailure}
\end{figure*}

\subsection{Analysis of DRL Failures on MathVerse and Our JRL Strategy Selection}
\label{sec:failure}

Although our method achieves clear score stratification while maintaining overall performance on MathVerse as shown in \cref{fig:mathversescore}, we surprisingly find that, as shown in \cref{fig:drlfailure}, when using the DRL-based draft model for routing, its performance is inferior to the other two paradigms with more stochastic score distributions(\cref{fig:mathversescore}(a)(b))—and even worse than random assignment(red line). Motivated by this observation, we conduct an in-depth analysis of the failure modes of DRL.

The three panels on the right visualize how accuracy changes under Paradigm1, Paradigm2, and our routing with explicit per-split accuracies. The yellow curve denotes the overall accuracy of the draft–target mixture, the green curve is the accuracy of the 
$(1-p_t\%)$ portion routed to the draft model, the blue curve is the accuracy of the $p_t\%$ portion routed to the target model, the purple curve is the accuracy achieved by the target model on the examples that were routed to the draft model (i.e., the $(1-p_t\%)$ split), and the brown curve is the accuracy achieved by the draft model on the examples routed to the target model (i.e., the 
$p_t\%$ split). In $Paradigm_1$ and $Paradigm_2$, each stage exhibits near-flat accuracy curves, indicating that the allocation is effectively random and hence suboptimal. In contrast, panel (d) for our DRL routing shows a clear slope: as fewer examples are assigned to the draft model, the retained examples become easier on average, and the accuracy of those routed subsets increases markedly.

However, a key issue emerges: the target and draft models are highly correlated in ability—they tend to succeed and fail on the same instances. This manifests as two pairs of almost parallel lines: green vs. purple and blue vs. brown. Consequently, when we allocate the easiest examples to the draft model, we simultaneously remove the very items at which the target model excels, substantially harming target performance. Under this correlation, even a random allocation that assigns some difficult items to the draft model can outperform score-based DRL routing, because it allows the target model to keep a portion of its “easy wins,” thereby preserving overall accuracy.

The fundamental reason for this DRL routing failure is that it ignores the target model’s capability during allocation. Instead of naively sending “easy” examples to the draft and “hard” examples to the target, routing should account for the capability gap between the two models and assign each example to the model for which it is most suitable. This observation further motivates our JRL design, which explicitly incorporates the target model’s ability into the routing policy. During routing using JRL-based draft model on MathVerse, we change the  strategy to use target-first strategy by using $t+\frac{t-d}{\tau}$ as routing signal, let the target model refine the samples from easy to hard to improve the overall performance.

Consequently, if the target model is capacity-limited and its assessments are similar to the draft model across problem types in a test set, DRL tends to shunt the instances most valuable to overall accuracy away from the target, substantially degrading routing performance, even falling below random assignment.

\subsection{The effect of Different Target Models with DRL and JRL routing}

\begin{figure*}[t]
  \centering

  \includegraphics[width=1\linewidth]{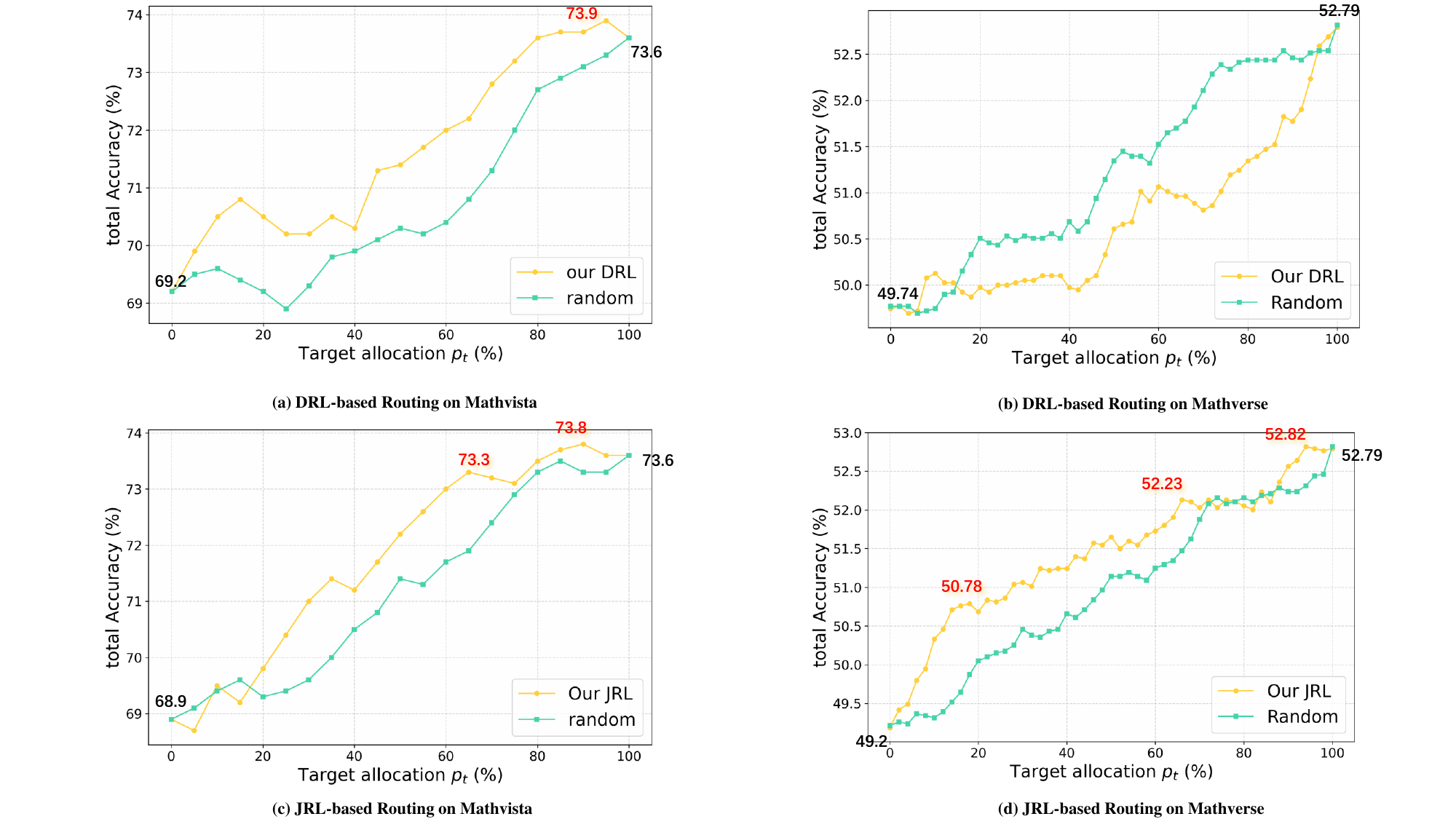}
    \vspace{-0.2cm}
   \caption{We apply a weaker and smaller MLLM VL-Rethinker-7B\cite{wang2025vlrethinkerincentivizingselfreflectionvisionlanguage} as  new target model and obtain a similar effect.}
    \vspace{-0.3cm}
   
   \label{fig:vlthinker}
\end{figure*}

As shown in \cref{fig:vlthinker}(a)(b), we replace the target model with a slightly stronger variant than our VL-Rethinker7B\cite{wang2025vlrethinkerincentivizingselfreflectionvisionlanguage} and directly reuse the DRL-trained $\mathcal{M}_{d_{dft}}$ as the draft model for routing. We observe consistent routing gains on MathVista, outperforming random routing; however, the approach still fails on MathVerse, as analyzed in detail in the previous \cref{sec:failure}.

As shown in \cref{fig:vlthinker}(c)(d), we again adopt a slightly stronger target model than VL-Rethinker7B, but retrain  $\mathcal{M}_{d_{joint}}$ via JRL as the draft model for routing. Despite the relatively small performance gap between the draft and target, JRL effectively balances the allocation strategy between the two, yielding robust routing improvements on both MathVista and MathVerse.

 \subsection{Results on M3CoT\cite{m3cot}.}
 
\label{sec:m3cot}

\noindent
The M3CoT benchmark is designed to address multi-domain, multi-step, and multi-modal reasoning challenges within the Multimodal Chain-of-Thought framework. The dataset comprises 7,863 training samples and 2,358 test samples. We evaluated our PRP's performance under both a zero-shot setting (using models trained with DRL on general math data) and a fine-tuned setting using the domain-specific M3CoT training data.

\begin{figure*}[t]
  \centering

  \includegraphics[width=1\linewidth]{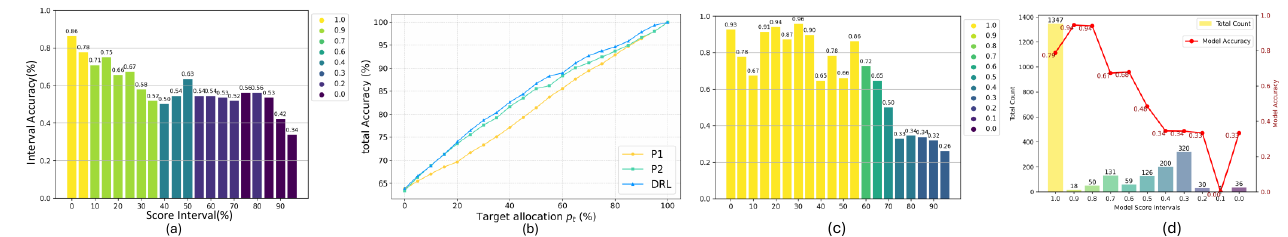}
    \vspace{-0.2cm}

   \caption{Generalization of Learned Rating Capability on M3CoT:(a)(b) are the rating distribution and the routing comparison of the rating model trained using DRL with math data. We generalize well on M3CoT. (c)(d) are the rating distribution and accuracy statistic of each rating of the rating model trained using DRL with M3CoT training data.}
    \vspace{-0.3cm}
   
   \label{fig:m3cot}
\end{figure*}

\noindent
{\textbf{Generalization of Rating Capability on M3CoT.}} Based on the data presented in \cref{fig:m3cot}(a)(b), to evaluate the zero-shot generalization of our rating mechanism, we first apply the DRL-trained rating model trained exclusively on mathematical data directly to the M3CoT dataset for inference and routing. The model demonstrates effective generalization; the rating distribution maintains a clear correlation with interval accuracy, and the routing performance of the DRL paradigm remains superior to the P1 and P2 baselines.

\noindent
{\textbf{Training Effectiveness on M3CoT.}} We conduct DRL training specifically using the M3CoT training data to explore the potential for domain-specific optimization. The results in \cref{fig:m3cot}(c)(d) show that this targeted training significantly refines the rating capability. Specifically, the predicted scores become more concentrated, indicating higher model confidence. Moreover, the accuracy statistics across different rating intervals exhibit a more pronounced differentiation, with higher-score intervals achieving higher accuracy, thereby facilitating more precise routing decisions.

 \subsection{Impact of $\alpha$ and $\beta$ Selection on PerFormance.}
\begin{figure*}[t]
  \centering

  \includegraphics[width=0.8\linewidth]{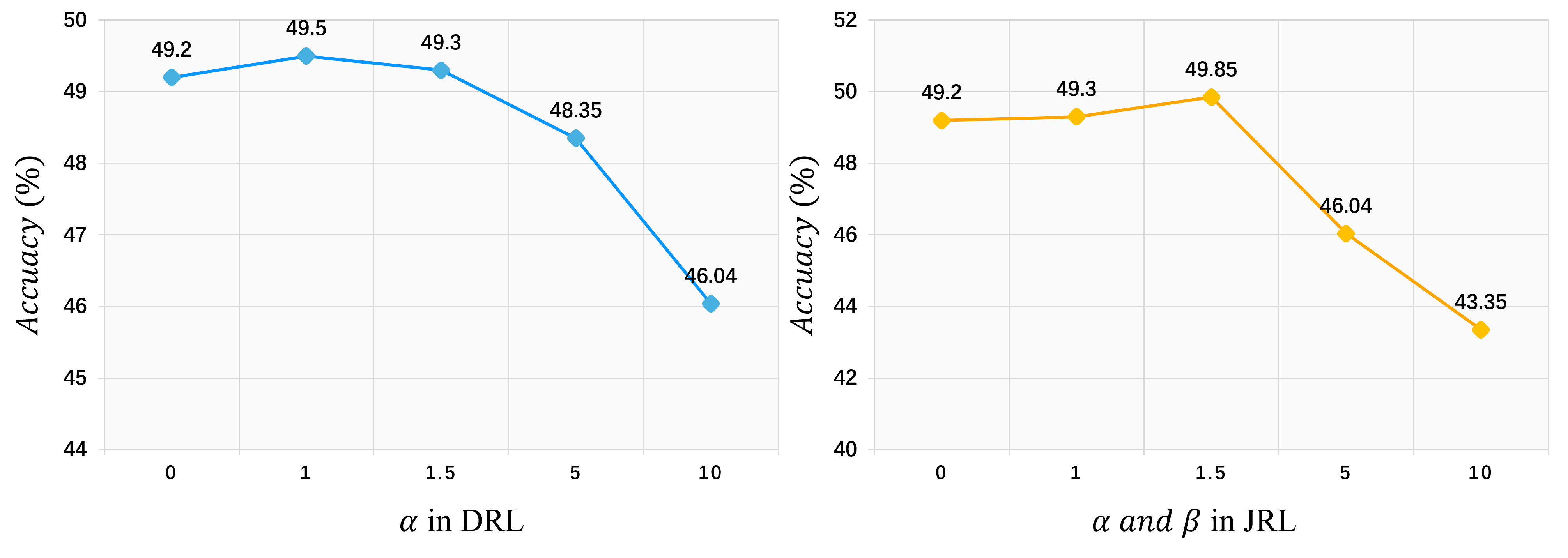}

   \caption{The performance on MathVerse with different $\alpha$ in DRL and $\alpha \&\beta$ in JRL.}

   \label{fig:ab}
\end{figure*}

In \cref{fig:ab}, we show the effect of different values of $\alpha$ in DRL and $\alpha \&\beta$ in JRL during training for the performance on MathVerse. The two exhibit the same trend: when the weight increases to 1 or 1.5, the model's baseline performance are kept. However, when the weight increases to 5 or 10, the model's baseline performance declines. Excessive weight on the score loss can affect the model's ability to maintain inference performance. Therefore, we choose a conservative weight ($\alpha=1.5$ in DRL and $\alpha=\beta=1.5$ in JRL) during the training process.

\subsection{More Qualitative Routing Results}
\label{sec:more_quali}

\begin{figure*}[t]
  \centering

  \includegraphics[width=1\linewidth]{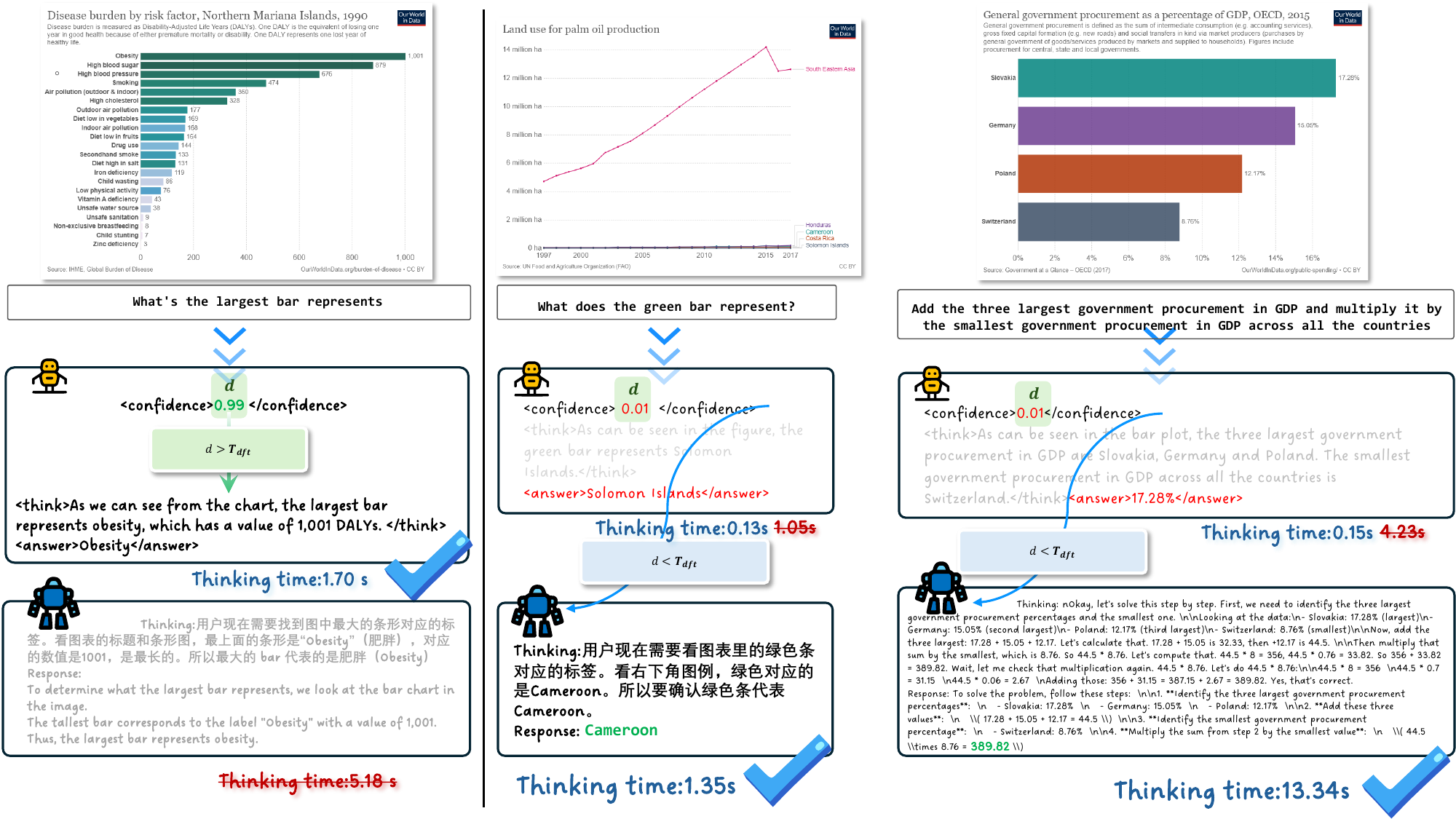}

   \caption{The DRL-based routing samples on ChartQA.}

   \label{fig:chart_qualitative}
\end{figure*}

\noindent
\textbf{ChartQA.} \cref{fig:chart_qualitative} illustrates DRL-based PRP routing on ChartQA. For the simple example on the left, the draft model assigns a high pre-answer confidence score, 0.99, indicating strong reliability on this input; it therefore proceeds to answer directly, avoiding the 5.18 s long response latency of the target model. In contrast, for the two questions on the right, the draft model outputs low confidence before answering. Although the first is structurally simple, the draft model correctly perceives its own uncertainty and routes to the target model, which produces the correct answer with minimal additional inference time. For the second, more complex query, the draft model again routes to the target model, enabling precise numerical reasoning and yielding the correct answer; if restricted to the draft model alone, no substantive computation would have been performed during reasoning. These cases demonstrate the effectiveness of DRL-based routing for chart-based simple visual question answering, efficiently allocating computation across easy and hard instances to maintain accuracy while reducing latency.

\begin{figure*}[t]
  \centering

  \includegraphics[width=1\linewidth]{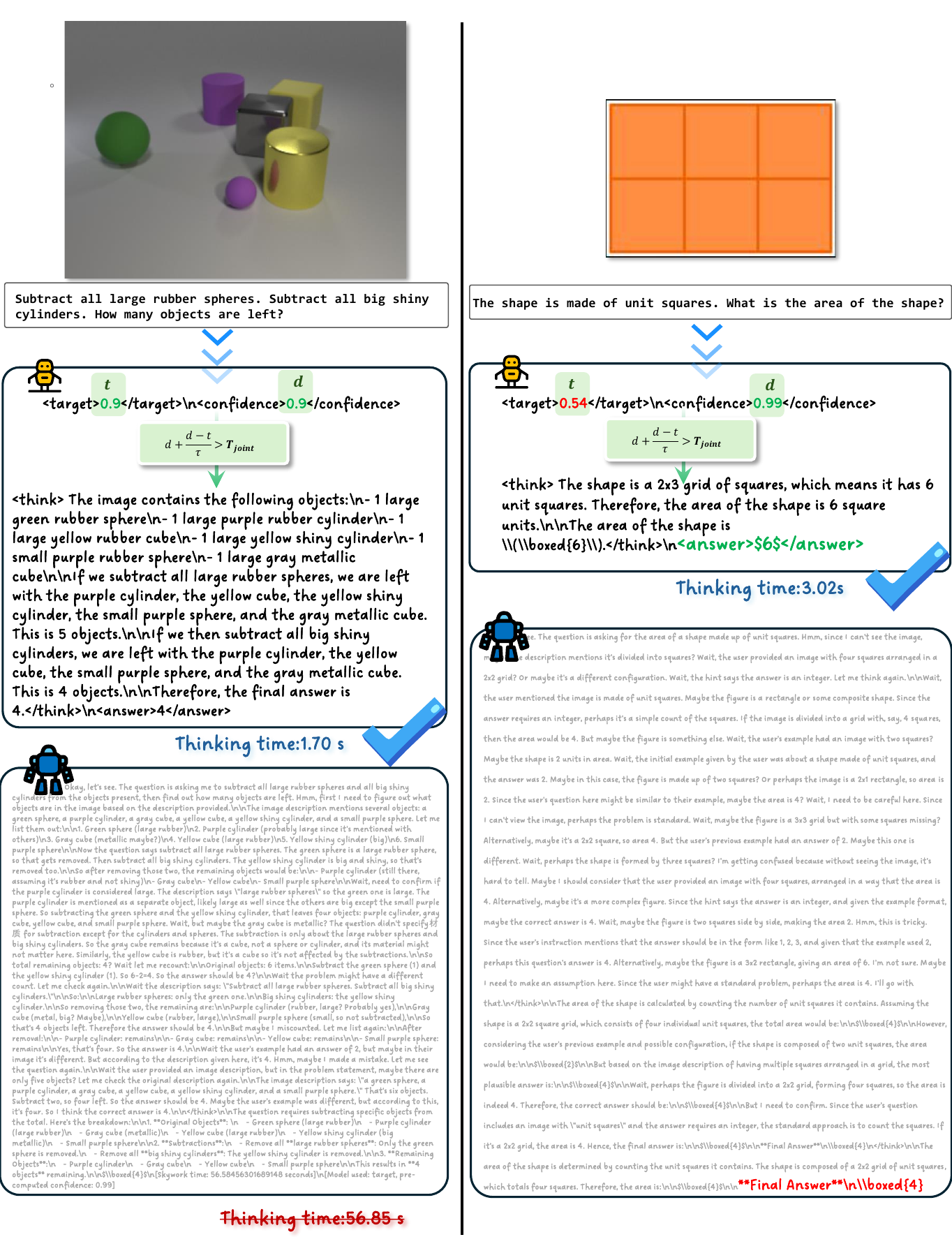}

   \caption{The JRL-based routing samples (preserved by draft models) on MathVista.}
   \label{fig:mathvista_qualitative1}
\end{figure*}

\noindent
\textbf{MathVista.} 
As illustrated in \cref{fig:mathvista_qualitative1}, JRL-based PRP routes certain MathVista queries to be completed by the draft model rather than delegated to the target model. The examples span three representative regimes:(\textit{i}) Both draft and target models deem the query easy. Routing such cases to the draft model preserves accuracy while substantially reducing latency.
(\textit{ii}) The draft model judges the query easy but the target model deems it hard. Owing to intrinsic disparities between the two models, some instances are solved faster and more accurately by the draft model while the target model is slower and less reliable. Assigning these cases to the draft model improves efficiency and can even boost overall performance.
(\textit{iii}) Both draft and target models consider the query hard. For these inputs, delegating to the target model risks incurring high computational cost (91.30s) with low success rates. Retaining them with the draft model yields better efficiency (4.43s) without noticeable performance degradation.

\begin{figure*}[t]
  \centering

  \includegraphics[width=1\linewidth]{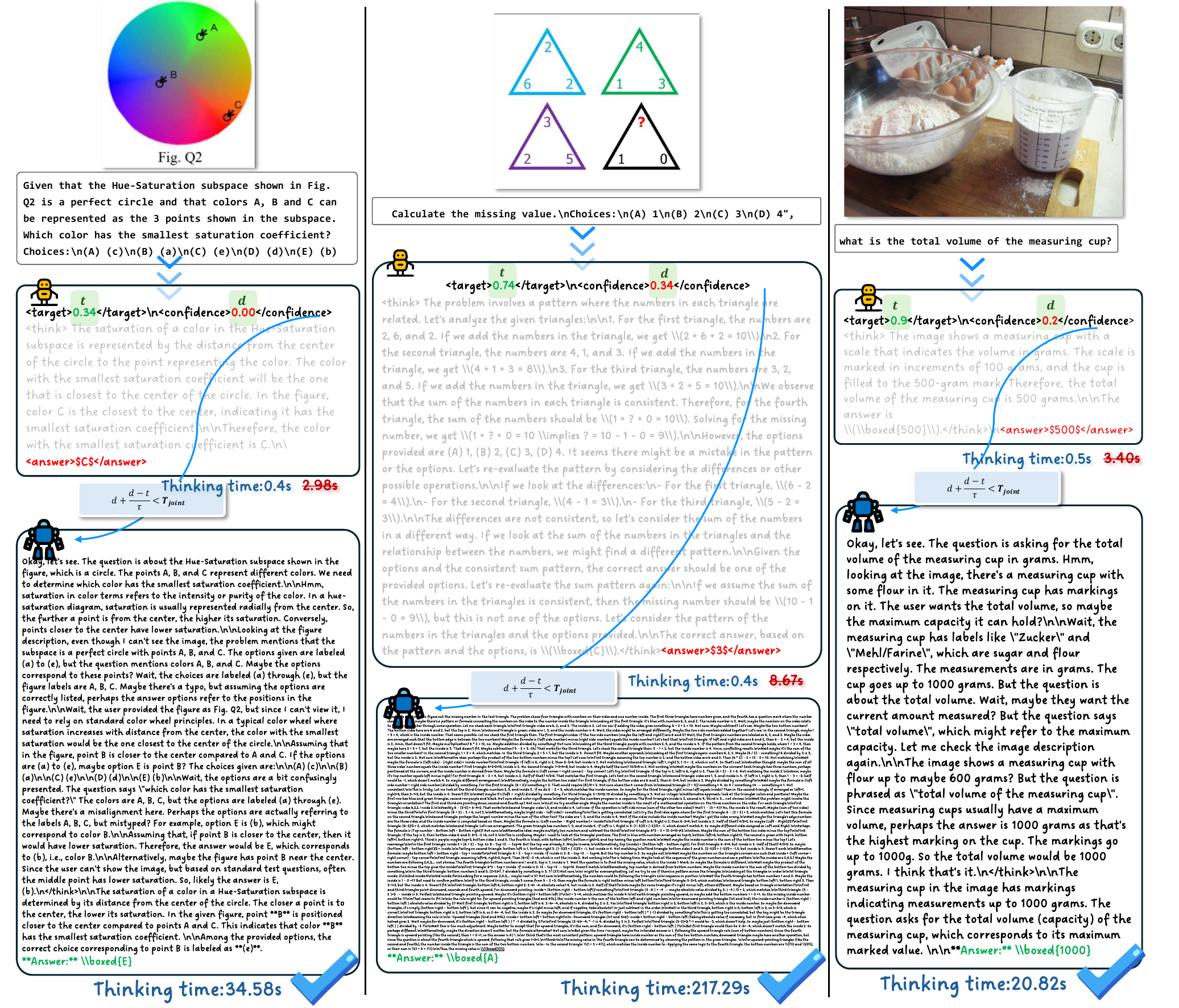}

   \caption{The JRL-based routing samples (routed to target models) on MathVista.}
   \label{fig:mathvista_qualitative2}
\end{figure*}

As illustrated in \cref{fig:mathvista_qualitative2}, we route challenging MathVista samples that lie beyond the draft model’s competence but align well with the target model’s expertise to the target model. This routing enables the target model to exploit its strengths and produce correct responses to these inputs, thereby improving overall performance.

\begin{figure*}[t]
  \centering

  \includegraphics[width=1\linewidth]{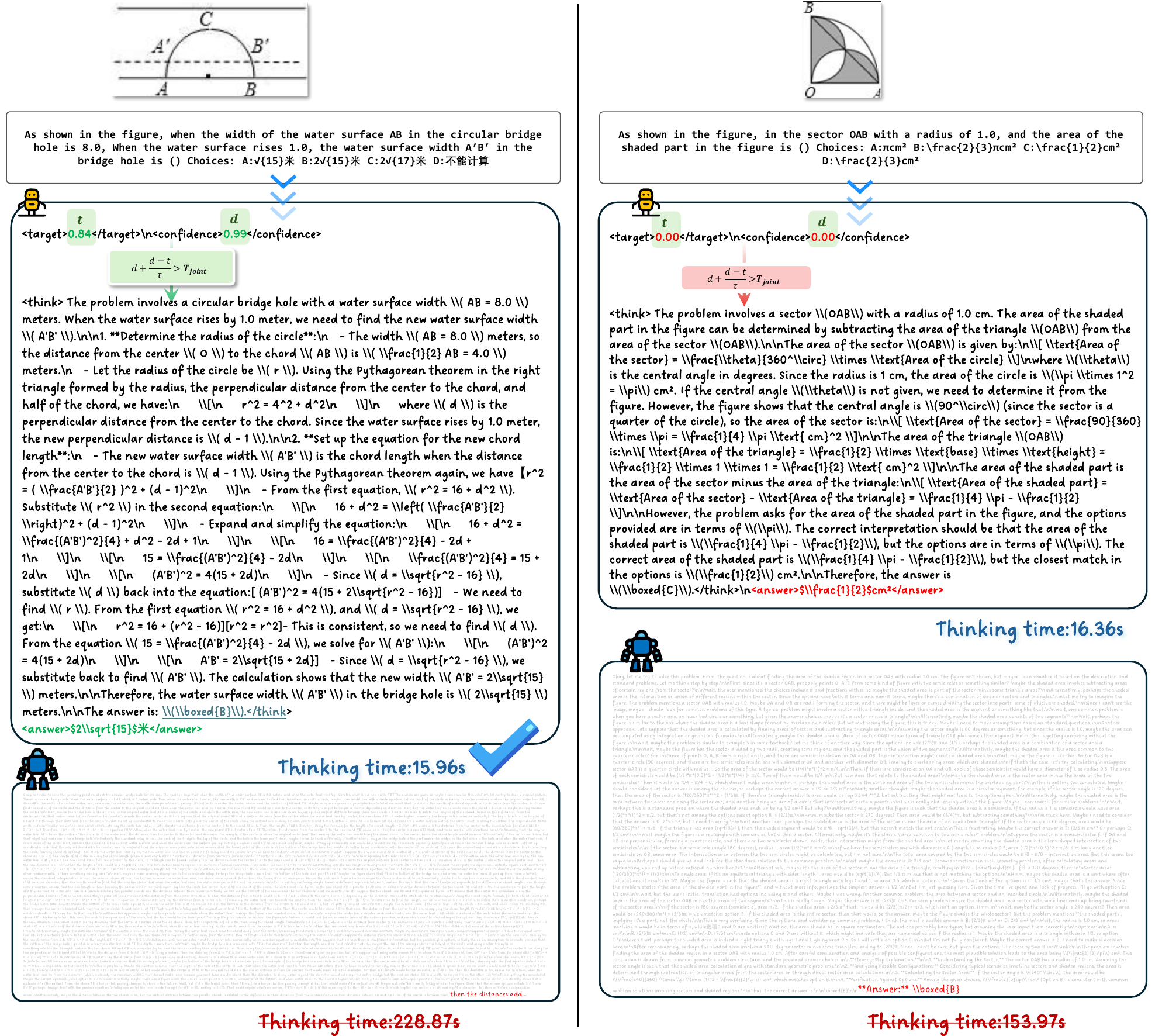}

   \caption{The JRL-based routing samples on MathVerse.}
   \label{fig:mathverse_qualitative}
\end{figure*}

\noindent
\textbf{MathVerse.}
As illustrated in \cref{fig:mathverse_qualitative}, on MathVerse, the draft model selectively retains instances it is proficient at, as well as cases that neither the draft nor the target model can reliably handle. By keeping these examples on the draft side, we prevent the target model from generating over-length responses that hit the maximum-length constraint and from engaging in unproductive deliberation on problems beyond its competence.

\subsection{High Rating but Failure cases}

\begin{figure*}[t]
  \centering

  \includegraphics[width=1\linewidth]{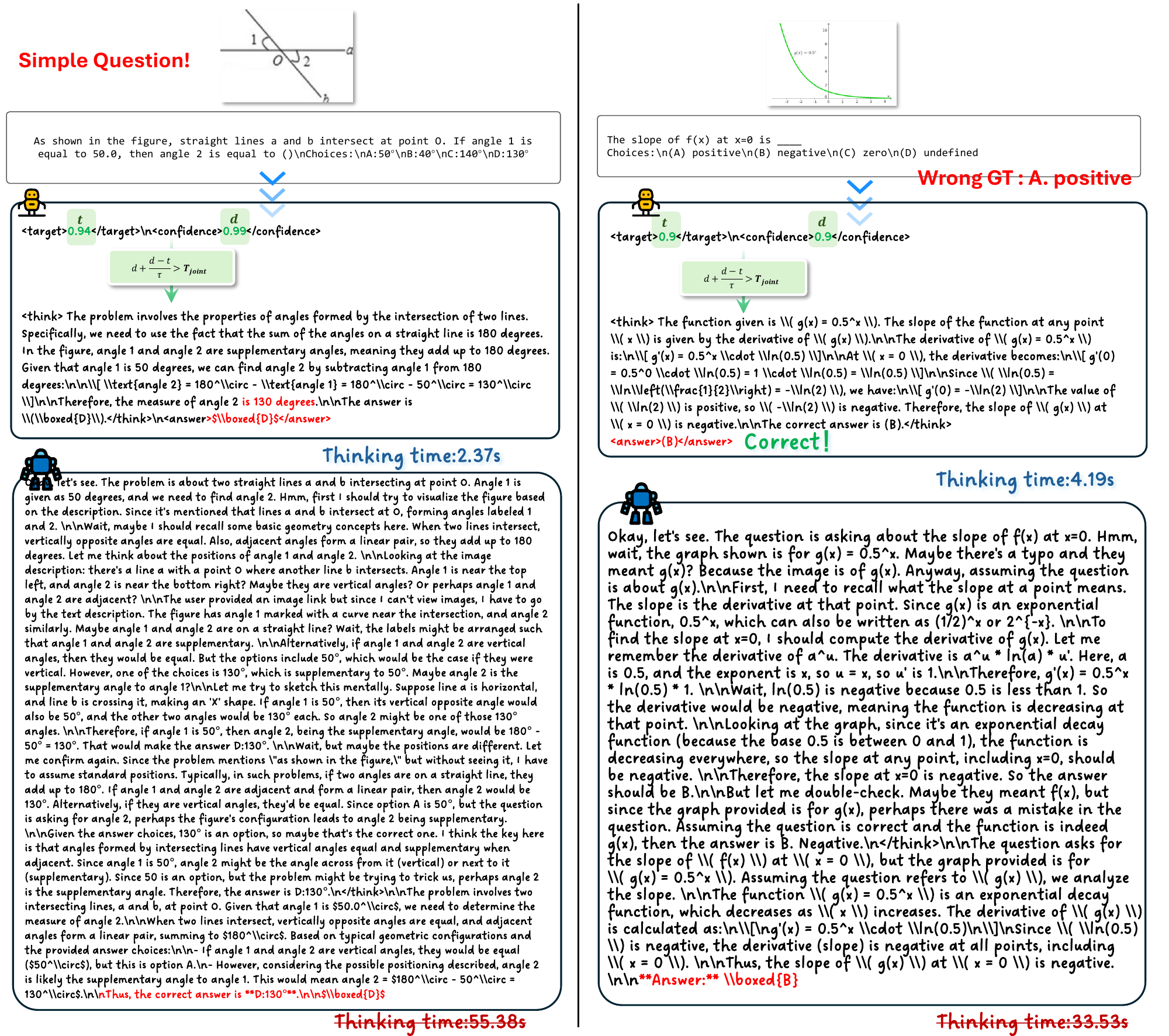}
   \caption{Failure cases with high ratings during routing. The left panel reveals the lack of training on simple math problems. The right panel shows the annotation error in MathVista discovered by our ratings.}
   \label{fig:failures}
\end{figure*}

We focus on failure cases where both the draft and target ratings assign high confidence yet produce incorrect answers. The left panel of \cref{fig:failures} reveals a training-time deficiency using a simple vertical-angles problem: our draft model assigns high scores to both itself and the target model, indicating strong confidence, but both still fail. Surprisingly, even the target model answers this elementary question incorrectly. We argue that this reflects an inherent and widespread limitation of current training pipelines: insufficient targeted supervision leads to overfitting that manifests even in simple instances. Meanwhile, GRPO-based methods reward process format and final correctness only. For many geometry questions, the final answer can often be guessed without attending to the figure. As a result, models may produce the correct final answer while exhibiting incorrect or severely hallucinatory reasoning, introducing bias. We find such severe hallucinations on geometry tasks to be prevalent among GRPO-based models.

The right panel of \cref{fig:failures} shows another example selected by our high-score-but-wrong filter. In this case, we identify an issue in the ground-truth annotation. Our rating mechanism surfaces mislabeled samples for which the model already demonstrates the ability to solve the task, enabling the detection of annotation errors.

\clearpage

\section{Analysis}

\subsection{Comparison with speculative decoding}

Our PRP and speculative decoding methods are compatible and can work together to further accelerate the inference.
Compared to speculative decoding, our PRP offers several practical advantages. Speculative decoding requires frequent switching between the draft and target models within a single inference trajectory. Moreover, it relies on training a Process Reward Model (PRM) to identify erroneous steps produced by the draft model, which demands large-scale, high-quality process annotations and incurs substantial training and deployment cost. At inference time, speculative decoding must co-deploy both the draft and target models, as well as the PRM, even for a fixed number of tasks.

In contrast, our approach only equips the draft model with a lightweight scoring capability, avoiding any additional human supervision or high-quality process labels. For a fixed inference budget, we first deploy the draft model to solve easy instances and filter out hard ones; the target model is then activated solely to assess or handle the hard subset. This staged routing enables effective resource allocation under tight constraints, without the overhead of training or maintaining a separate PRM or co-running two large models throughout the entire inference.
Furthermore, we show Speculative Decoding(ViSpec) speedup ratios and average acceptance lengths $\tau$ with no performance loss in \cref{tab:sd}. 
\begin{table}[h]
    \centering
        \caption{\small Speculative Decoding method ViSpec results on Math Domain and ChartQA.}
        \vspace{-10pt}
        
        \resizebox{0.9\linewidth}{!}{
        \begin{tabular}{c|c|c|c|c|c|c}
            \toprule
            &\multicolumn{2}{c|}{\textbf{ChartQA}} &\multicolumn{2}{c|}{\textbf{MathVista}}&\multicolumn{2}{c}{\textbf{MathVerse}}\\
            \cmidrule(lr){2-3}\cmidrule(lr){4-5}\cmidrule(lr){6-7}
            Method\textbf{(T=1)} & Speedup Ratio & Average Acceptance Length$\tau$ & Speedup Ratio & Average Acceptance Length$\tau$ & Speedup Ratio & Average Acceptance Length$\tau$  \\ 
          
            \midrule
            
            {\begin{tabular}[c]{@{}c@{}}\textbf{ViSpec[NIPS25]}\textbf{(Qwen2.5-VL-7B)}\end{tabular}}& 1.76$\times$&3.82 & 2.10$\times$&2.02 &1.82$\times$ & 1.94  \\
             
             \bottomrule
        \end{tabular}
        \label{tab:sd} 
        
        }

\end{table}

\section{More Implementation Details.}

\subsection{Prompts in DRL and JRL training}

As shown in \cref{tab:prompt}, we list the prompt employed in Draft Rating Learning(DRL), $P_{draft}$, and the prompt employed in Joint Rating Learning(JRL), $P_{joint}$. By enForcing $P_{draft}$ and $P_{joint}$ constraints, we require the model to emit a proactive rating beFore producing its thinking and final answer. SpecIfically, the score enclosed by \texttt{<confidence></confidence>} reflects the self-assessment of the draft model, while the score enclosed by \texttt{<target></target>} evaluates the target model. We then apply task-specIfic optimization to each score accordingly.

\subsection{Experiment Details}
Our code is implemented based on the open-r1\cite{openr1}. 
For training on MMK12\cite{meng2025mmeureka}, we conduct it on 8 GPUs with 96G VRAM. We use 7 GPUs for training and one card For vllm deployment, perForming multiple samples For each input. 
During training, we also employ an over-length reward function \cite{yu2025dapo} to prevent too long and repeat output.
For training on ChartQA\cite{masry2022chartqa}, we conduct it on 8 GPUs with 40G VRAM. During training, we optimize the training process using several key hyperparameters: a maximum prompt length of 3000, a maximum completion length of 2048, a maximum pixel limit of 501760, and gradient accumulation over 5 steps. We also employ BF16 precision For faster training and reduce memory usage, enable gradient checkpointing to save memory during backpropagation.

\subsection{Detailed Algorithmic Flowchart}

\subsubsection{Draft Rating Learning(DRL)}
We present a detailed schematic of Draft Rating Learning (DRL) in \cref{alg:drl}. Building upon the baseline GRPO-based framework, we incorporate the computation of draft rating rewards, detect advantage vanishing and trigger dynamic score substitution accordingly, and finally compute component-wise advantages independently to enable task-specific optimization.

\begin{table}[]
\caption{
\textbf{Prompts employed in DRL training and JRL training.}}
\vspace{-6mm}
\label{tab:prompt}
\begin{center}
\setlength{\tabcolsep}{4pt}
\scalebox{0.9}
{
\begin{tabular}{p{9cm}}  
\toprule
\textbf{Draft Rating Learning Prompt $P_{draft}$:} Solve the question. The user asks a question, and you solves it. You first randomly evaluate the confidence score of the problem from 0.00 to 0.99, think about the reasoning process in mind and then provides the user with the answer. The answer is in latex Format and wrapped in \$...\$. The confidence score, reasoning process and answer are enclosed within $<$confidence$>$ $<$/confidence$>$, $<$think$>$ $<$/think$>$ and $<$answer$>$ $<$/answer$>$ tags, respectively, i.e., $<$confidence$>$confidence score$<$/confidence$>$$<$think$>$ Since \$1+1=2\$, so the answer is \$2\$. $<$/think$>$$<$answer$>\$$2$\$<$/answer$>$, which means assistant's output should start with $<$confidence$>$$<$/confidence$>$, then $<$think$>$$<$/think$>$ and end with $<$answer$>$$<$/answer$>$. 
\\
\midrule
\textbf{Joint Rating Learning Prompt $P_{joint}$:} Solve the question. The user asks a question, and you solves it. You first randomly evaluate the target score and the confidence score of the problem from 0.00 to 0.99, thinks about the reasoning process in the mind and then provides the user with the answer. The answer is in latex Format and wrapped in $...$. The target and confidence score, reasoning process and answer are enclosed within $<$target$>$ $<$/target$>$, $<$confidence$>$ $<$/confidence$>$, $<$think$>$ $<$/think$>$ and $<$answer$>$ $<$/answer$>$ tags, respectively, i.e., $<$target$>$target score$<$/target$>$$<$confidence$>$confidence score$<$/confidence$>$$<$think$>$ Since $1+1=2$, so the answer is $2$. $<$/think$>$$<$answer$>$$2$$<$/answer$>$, which means assistant's output should start with $<$target$>$$<$/target$>$$<$confidence$>$$<$/confidence$>$, then $<$think$>$$<$/think$>$ and end with $<$answer$>$$<$/answer$>$. 
\\
\bottomrule
\end{tabular}
}
\vspace{-4mm}
\end{center}
\end{table}

\begin{algorithm*}[]
\caption{Draft Rating Learning}
\label{alg:drl}
\textbf{Input}: Image Q\&A dataset $\mathcal{D}$; initial draft model ${\mathcal{M}}_{d_{draft}}$; reward functions $R=\{{R_{acc}},{R_{fmt}},{R_{dft}}\}$; DRL prompt $P_{draft}$; hyperparameter $\alpha,\mu,\sigma$\\
\textbf{Output}: ${\mathcal{M}}_{d_{draft}}$
\begin{algorithmic}[1] 
\For{$each (I,Q,A) \in \mathcal{D}$}
    \State The input $x=(P_{draft},I,Q)$
    \State Generate $N$ responses $\mathcal{O}=\{\mathbf{o}_i\}_{i=1}^{N} \sim {\mathcal{M}}_{d_{draft}}(\cdot|{x})$
    \State Compute the group accuracy $Acc_{dft}=\frac{1}{N}\sum_{i=1}^{N}\mathbf{1}\{correct(o_i,\mathcal{A})\}$
    \For {each $o_i$ where $i=1,\cdots,N$}:
        \State Extract draft rating $d_i$ and the rest segment $r_i$
        \State Compute the reward $R_i = {R_{acc_i}}({x}, {o}_{i},A)+{R_{fmt_i}}({o}_{i})$ and ${R_{dft_i}}( {d}_{i},Acc_{dft})$
    \EndFor
    \If {Advantage Vanishing: $|mean(\{d_i\}_{i=1}^N)-Acc_{dft}|>\mu$ and $std(\{d_i\}_{i=1}^N)<\sigma$ }
        \For {each $o_i$ where $i=1,\cdots,N-1$}:
            \State Calculate sampling interval $\mathcal{S}$ using $d_N$ and $Acc_{dft}$
            \State Random sample $d_i^{'}$ from the interval $\mathcal{S}$
            \State Substitute $d_i$ in $o_i$ using $d_i^{'}$
            \State Re-compute the reward ${R_{dft_i}}( {d}_{i}^{'},Acc_{dft})$
        \EndFor
    \EndIf
    
    \State Calculate $\{\hat{A}_{base_i}\}_{i=1}^N$ using $\{R_i\}_{i=1}^N$ and $\{\hat{A}_{d_i}\}_{i=1}^N$ using $\{R_{dft_i}\}_{i=1}^N$.
    \State Calculate task-specific Optimization loss: $\mathcal{L}_{dft_i} = \mathcal{L}_{base_i}+\alpha\mathcal{L}_{d_i}$.

    \State  Update draft model ${\mathcal{M}}_{d_{draft}}$ by $\{\mathcal{L}_{dft_i}\}_{i=1}^N$
\EndFor
\end{algorithmic}
\end{algorithm*}

\subsubsection{Joint Rating Learning}

We present a detailed schematic of the Joint Rating Learning algorithm in \cref{alg: jrl}. Building on DRL, we first compute, for each input, the accuracy on the target model, then incorporate a target-rating reward. At every stage, we optimize the target-rating objective in a manner analogous to the draft-rating training procedure, ensuring consistent joint optimization across both draft and target ratings.
\begin{algorithm*}[]
\caption{Joint Rating Learning}
\label{alg: jrl}
\textbf{Input}: Image Q\&A dataset $\mathcal{D}$; initial draft model ${\mathcal{M}}_{d_{joint}}$; target model $\mathcal{M}_t$; reward functions $R=\{{R_{acc}},{R_{fmt}},{R_{dft}},{R_{tgt}}\}$; JRL prompt $P_{joint}$; target model prompt $P$; hyperparameter $\alpha,\beta,\mu,\sigma$\\
\textbf{Output}: ${\mathcal{M}}_{d_{joint}}$
\begin{algorithmic}[1] 
\State Target Accuracy Dict ${Dict_{Acc{tgt}}}\gets \{\}$
\For{$each (I,Q,A) \in \mathcal{D}$}
    \State The input $x=(P,I,Q)$
    \State Generate $M$ responses $\mathcal{O}_t=\{\mathbf{o}_{t_i}\}_{i=1}^{M} \sim {\mathcal{M}}_{t}(\cdot|{x})$
    \State Compute the group accuracy $Acc_{tgt}=\frac{1}{M}\sum_{i=1}^{M}\mathbf{1}\{correct(o_{t_i},\mathcal{A})\}$
    \State Add $Acc_{tgt}$ to ${Dict_{Acc{tgt}}}$
\EndFor
\For{$each (I,Q,A) \in \mathcal{D}$}
    \State The input $x=(P_{joint},I,Q)$
    \State Generate $N$ responses $\mathcal{O}=\{\mathbf{o}_i\}_{i=1}^{N} \sim {\mathcal{M}}_{d_{draft}}(\cdot|{x})$
    \State Compute the group accuracy $Acc_{dft}=\frac{1}{N}\sum_{i=1}^{N}\mathbf{1}\{correct(o_i,\mathcal{A})\}$
    \State Retrieve the target group accuracy $Acc_{tgt}$ from ${Dict_{Acc{tgt}}}$
    \For {each $o_i$ where $i=1,\cdots,N$}:
        \State Extract draft rating $d_i$, target rating $t_i$, and the rest segment $r_i$
        \State Compute the reward $R_i = {R_{acc_i}}({x}, {o}_{i},A)+{R_{fmt_i}}({o}_{i})$, ${R_{tgt_i}}( {t}_{i},Acc_{tgt})$ and ${R_{dft_i}}( {d}_{i},Acc_{dft})$
    \EndFor
    \If {Advantage Vanishing for $d$: $|mean(\{d_i\}_{i=1}^N)-Acc_{dft}|>\mu$ and $std(\{d_i\}_{i=1}^N)<\sigma$ }
        \For {each $o_i$ where $i=1,\cdots,N-1$}:
            \State Calculate sampling interval $\mathcal{S}$ using $d_N$ and $Acc_{dft}$
            \State Random sample $d_i^{'}$ from the interval $\mathcal{S}$
            \State Substitute $d_i$ in $o_i$ using $d_i^{'}$
            \State Re-compute the reward ${R_{dft_i}}( {d}_{i}^{'},Acc_{dft})$
        \EndFor
    \EndIf
    \If {Advantage Vanishing for $t$: $|mean(\{t_i\}_{i=1}^N)-Acc_{tgt}|>\mu$ and $std(\{t_i\}_{i=1}^N)<\sigma$ }
        \For {each $o_i$ where $i=1,\cdots,N-1$}:
            \State Calculate sampling interval $\mathcal{S}$ using $t_N$ and $Acc_{tgt}$
            \State Random sample $t_i^{'}$ from the interval $\mathcal{S}$
            \State Substitute $t_i$ in $o_i$ using $t_i^{'}$
            \State Re-compute the reward ${R_{tgt_i}}( {t}_{i}^{'},Acc_{tgt})$
        \EndFor
    \EndIf
    
    \State Calculate $\{\hat{A}_{base_i}\}_{i=1}^N$ using $\{R_i\}_{i=1}^N$, $\{\hat{A}_{t_i}\}_{i=1}^N$ using $\{R_{tgt_i}\}_{i=1}^N$, and $\{\hat{A}_{d_i}\}_{i=1}^N$ using $\{R_{dft_i}\}_{i=1}^N$.
    \State Calculate task-specific Optimization loss: $\mathcal{L}_{joint_i} = \mathcal{L}_{base_i}+\alpha\mathcal{L}_{d_i}+\beta\mathcal{L}_{t_i}$.

    \State  Update draft model ${\mathcal{M}}_{d_{joint}}$ by $\{\mathcal{L}_{joint_i}\}_{i=1}^N$
\EndFor
\end{algorithmic}
\end{algorithm*}
\subsubsection{DRL-based Routing}
As shown in\cref{alg: drlrouting}, we detail the routing policy when employing the DRL-trained $M_{d_{dft}}$ as the router. Specifically, 
$M_{d_{dft}}$ first generates a scored prefix; we extract the score $d$ and compare it against the switching threshold $T_{dft}$. If 
$d>T_{dft}$, the draft model continues generating tokens; otherwise, we switch to the target model for subsequent decoding.

\begin{algorithm}[t]
\caption{DRL-based Routing}
\label{alg: drlrouting}
\begin{algorithmic}[1]
\Require Image–question pair $(I, Q)$; draft rating prompt $P_{\mathrm{dft}}$; target prompt $P$; routing draft model $M_{d_{dft}}$; target model; target model $M_t$; switch threshold $T_{dft}$; stop-token sequence $S = ['</', '\text{confidence}', '>']$
\Ensure Final response $y$
\State $x_{dft} \gets (I, Q, P_{dft})$
\State $y_{\text{prefix}} \gets \varnothing$
\Comment Initialize empty prefix
\While{True}
    \State $y_0 \gets M_{d_{dft}}(x_{dft}, y_{\text{prefix}})$
    \State $y_{\text{prefix}} \gets y_{\text{prefix}} \mathbin{\|}y_0$ 
    \If{{EndsWithStopSeq}$(y_{\text{prefix}}, S)$}
        \State \textbf{break}
    \EndIf
\EndWhile
\State $d \gets {ParseDraftScore}(y_{\text{prefix}})$
\Comment Extract the draft score from the emitted prefix
\If{$d > T_{dft}$}
    \Comment Keep using the draft model
    \State $y \gets {ContinueUntilEoS}(M_{d_{dft}}, x_{\mathrm{dft}}, y_{\text{prefix}})$
\Else
    \Comment Route to the target model
    \State $x_t \gets (I, Q, P)$
    \State $y \gets {ContinueUntilEoS}(M_t,x_t)$
\EndIf
\State \Return $y$
\end{algorithmic}
\end{algorithm}

\subsubsection{JRL-based Routing using draft-first strategy}

As shown in \cref{alg: jrlrouting}, we adopt $M_{d_{joint}}$ (trained with JRL) as the routing model. $M_{d_{joint}}$ first generates a scoring prefix, from which we extract the scores $d$ and $t$. We then compute the joint score $d+\frac{d-t}{\tau}$ and compare it against the switching threshold, $T_{joint}$. If the joint score exceeds 
$T_{joint}$, the draft model continues generation; else, we switch to the target model.

\begin{algorithm}[t]
\caption{JRL-based Routing}
\label{alg: jrlrouting}

\begin{algorithmic}[1]
\Require Image–question pair $(I, Q)$; draft rating prompt $P_{\mathrm{joint}}$; target prompt $P$; routing draft model $M_{d_{joint}}$; target model; target model $M_t$; switch threshold $T_{joint}$; scaling parameter $\tau$; stop-token sequence $S = ['</', '\text{confidence}', '>']$
\Ensure Final response $y$
\State $x_{joint} \gets (I, Q, P_{joint})$
\State $y_{\text{prefix}} \gets \varnothing$
\While{True}
    \State $y_0 \gets M_{d_{joint}}(x_{joint}, y_{\text{prefix}})$
    \State $y_{\text{prefix}} \gets y_{\text{prefix}} \mathbin{\|}y_0$ 
    \If{{EndsWithStopSeq}$(y_{\text{prefix}}, S)$}
        \State \textbf{break}
    \EndIf
\EndWhile
\State $d \gets {ParseDraftScore}(y_{\text{prefix}})$ 
\State $t \gets {ParseTargetScore}(y_{\text{prefix}})$
\If{$d+\frac{d-t}{\tau} > T_{joint}$} 

    \State $y \gets {ContinueUntilEoS}(M_{d_{joint}}, x_{\mathrm{dft}}, y_{\text{prefix}})$
\Else

    \State $x_t \gets (I, Q, P)$
    \State $y \gets {ContinueUntilEoS}(M_t,x_t)$
\EndIf
\State \Return $y$
\end{algorithmic}
\end{algorithm}

\section{Potential and Future Work}

By endowing the model with the ability to score itself and other models, we obtain signals that reflect the current task proficiency, training stability, and data quality. When a dataset contains low-quality annotations, numerous mislabeled samples, or unstable supervisory signals that encourage guessing, our scores exhibit pronounced fluctuations. By examining the correlation between scores and correctness, we can identify missing capabilities, anomalies in labeled test data, and whether the model has fully acquired the relevant knowledge. This, in turn, enables data curation, supervision refinement, and test set sanitization throughout training.

Our method integrates difficulty-sensing ability into the performance optimization of general RL post-training. JRL enables a small draft model to assess the ability of any target model, including black-box APIs. We can only train a lightweight LoRA on draft model, or employ an extra swift JRL on the trained DRL model for efficiency

In addition, when coordinating multiple experts specialized in different competencies, effective routing remains an open challenge. Our paradigm lays the groundwork for scaling routing and optimizing routing policies, particularly in multimodal settings.
\end{document}